\documentclass[twocolumn, switch]{article} 

\usepackage{preprint}

\usepackage{amsmath, amsthm, amssymb, amsfonts}

\usepackage[numbers,square]{natbib}
\usepackage{natbib}

\usepackage[utf8]{inputenc}	
\usepackage[T1]{fontenc}	
\usepackage{xcolor}		
\usepackage[colorlinks = true,
            linkcolor = purple,
            urlcolor  = blue,
            citecolor = cyan,
            anchorcolor = black]{hyperref}	
\usepackage{booktabs} 		
\usepackage{nicefrac}		
\usepackage{microtype}		
\usepackage{lineno}		
\usepackage{float}			

\usepackage{newfloat}
\DeclareFloatingEnvironment[name={Supplementary Figure}]{suppfigure}
\usepackage{sidecap}
\sidecaptionvpos{figure}{c}

\usepackage{titlesec}
\titlespacing\section{0pt}{12pt plus 3pt minus 3pt}{1pt plus 1pt minus 1pt}
\titlespacing\subsection{0pt}{10pt plus 3pt minus 3pt}{1pt plus 1pt minus 1pt}
\titlespacing\subsubsection{0pt}{8pt plus 3pt minus 3pt}{1pt plus 1pt minus 1pt}

\title{Towards an Automatic System for Extracting Planar Orientations from Software Generated Point Clouds}

\usepackage{xwatermark}
\newwatermark[firstpage,color=gray!60,angle=90,scale=0.32, xpos=-4.05in,ypos=0]{\href{https://doi.org/}{\color{gray}{}}}
\newwatermark[firstpage,color=gray!60,angle=90,scale=0.32, xpos=3.9in,ypos=0]{\href{https://doi.org/}{\color{gray}{}}}
\newwatermark[firstpage,color=gray!90,angle=0,scale=0.28, xpos=0in,ypos=-5in]{*correspondence: \texttt{jkissiam@uwo.ca}}

\usepackage{authblk}

\author[1, 3\thanks{\tt{jkissiam@uwo.ca}}]{J. Kissi-Ameyaw}
\author[1, 3]{K. McIsaac}
\author[1]{X. Wang}
\author[2, 3]{G. R. Osinski}
\affil[1]{Department of Electrical and Computer Engineering, Thompson Engineering Building, Western University, London, Ontario, Canada, N6A 5B9}
\affil[2]{Department of Earth Sciences, Western University, 1151 Richmond Street N., London, Ontario, Canada, N6A 5B7}
\affil[3]{Institute for Earth and Space Exploration, Western University, 1151 Richmond Street, London, Ontario, Canada, N6A 3K7}

\usepackage{graphicx}
\usepackage{todonotes}
\usepackage{tensor}
\usepackage{parskip}
\usepackage{wrapfig}
\usepackage{subcaption}

\usepackage{algpseudocode}
\usepackage{algorithm}

\begin{document}

\twocolumn[ 
  \begin{@twocolumnfalse} 
  
\maketitle

\begin{abstract}
In geology, a key activity is the characterisation of geological structures (surface formation topology and rock units) using Planar Orientation measurements such as Strike, Dip and Dip Direction. In general these measurements are collected manually using basic equipment; usually a compass/clinometer and a backboard, recorded on a map by hand. Various computing techniques and technologies, such as Lidar, have been utilised in order to automate this process and update the collection paradigm for these types of measurements. Techniques such as Structure from Motion (SfM) reconstruct of scenes and objects by generating a point cloud from input images, with detailed reconstruction possible on the decimetre scale. SfM-type techniques provide advantages in areas of cost and usability in more varied environmental conditions, while sacrificing the extreme levels of data fidelity. Here is presented a methodology of data acquisition and a Machine Learning-based software system: \textbf{GeoStructure}, developed to automate the measurement of orientation measurements. Rather than deriving measurements using a method applied to the input images, such as the Hough Transform, this method takes measurements directly from the reconstructed point cloud surfaces. Point cloud noise is mitigated using a Mahalanobis distance implementation. Significant structure is characterised using a k-nearest neighbour region growing algorithm, and final surface orientations are quantified using the plane, and normal direction cosines.
\end{abstract}
\vspace{0.35cm}

  \end{@twocolumnfalse} 
] 


\section{Introduction}

Characterisation of surface geological structures using orientation measurements is a fundamental part of the geologists' tool set~\cite{Fosson2016,Sirat2001,Feng2001,Post2001,Kemeny2003,Olaniyan2014,Adam2000, Hecht2008,Chen2015,Rousell2003a,Tuchscherer2002a,Boerner2000a,Wood1998}. There are two main classes of "structure" in geology: planar and linear. Planar structures include things like bedding planes and fault surfaces, while linear structure consist mainly of lineations~\cite{Fosson2016}. Surface and underlying structure is inferred by measuring Planar formations using Strike, Dip and Dip Direction. Hypotheses are constructed based on the relationships between characterised areas.

The traditionally manual collection of these measurements makes heavy demands of the collector in time, expense, precluding spontaneity in field work, and, coverage is limited to areas that can be physically reached. In general these measurements are collected  using basic equipment; usually a compass/clinometer and a backboard and hand recorded. Factor in that these structures are themselves often many, many times larger than the data collector, and it becomes apparent how these considerations dramatically impact the level of data acquisition possible on each target site.

With Planar Orientation measurements being one of primary tools in the characterisation of surface features, much work has been done with the aim of augmenting these processes to reduce time, resources, safety hazards and increase coverage in collection~\cite{Sirat2001,Haid2016,Feng2001, Mah2013,Vasuki2014,Kemeny2003,Post2001,Bellian2005, Tavani2014, Chen2015}.

The aim of this work is to demonstrate a system that can provide an accurate, automated characterisation of PO measurements. The idea is to measure the data structure directly, instead of inferring the 3D dimensions from sample images. This process requires numerical techniques that must necessarily make large [mathematical] assumptions about the geometry being examined~\cite{Kemeny2003,Post2001}, before providing measurement estimations. In order utilise the MVG in this way, a specific data collection strategy is developed and integrated into the process, so that fidelity of the software generated point cloud can be maximised, and, highest accuracy of the extracted PO obtained. This stage uses pre-existing software and tools to collect the sample data, generate the point cloud and perform pre-processing.

The final stage of the process involves the development of ML software, GeoStructure, to remove any remaining outliers and process the point cloud into surfaces using a nearest-neighbour region growing technique. Once categorised into macro regions, PO measures are extracted and compared to pre-acquired ground truth measurements, using a dimensionless accuracy metric. Further work is done, regarding improving visualisation, to attempt to optimise convenience for the non-technical user.

\begin{figure*}[h]
	\centering
	\includegraphics[scale=.60]{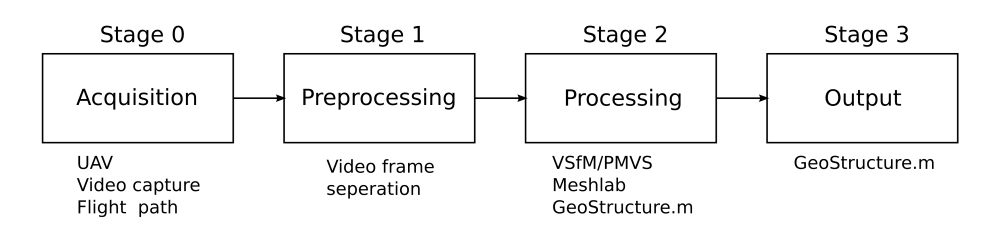}
	\caption[Software Toolchain Workflow]{Overall work flow of the proposed system.}
	\label{fig:toolchain dataflow}
\end{figure*}

\section{Related Work}
Although this is one of the most basic activities a geologist performs in the field, when broken down, the collection of these measurements is a complex process. Therefore, to create a system for automatic extraction of PO's requires the amalgamation of concepts from several areas. Many of the current approaches to this problem consist of image analysis techniques utilising various types of transformations, such as the Hough Transform, to infer 3D characteristics~\cite{Post2001, Kemeny2003,Vosselman2004a,Vasuki2014,Brown2002,Brown2005,Lowe1999a,Lowe2004} but the aim of this work is to measure the point cloud directly, as the collector would measure the rock with their instrument.

\subsection{Automating Orientation Extraction}

Work by Gigli et al.~\cite{Gigli2011} in this area, resulted in an integrated software to semi-automatically analyse and derive orientation measurements from the Lidar point clouds of geological formations. This was achieved using voxelisation to evaluate the point space and POs were derived using a direction cosine methodology \label{direction_cosines} similar in implementation to Feng et al.~\cite{Feng2001}.

\begin{figure}[h]
	\centering
	\includegraphics[scale=.09]{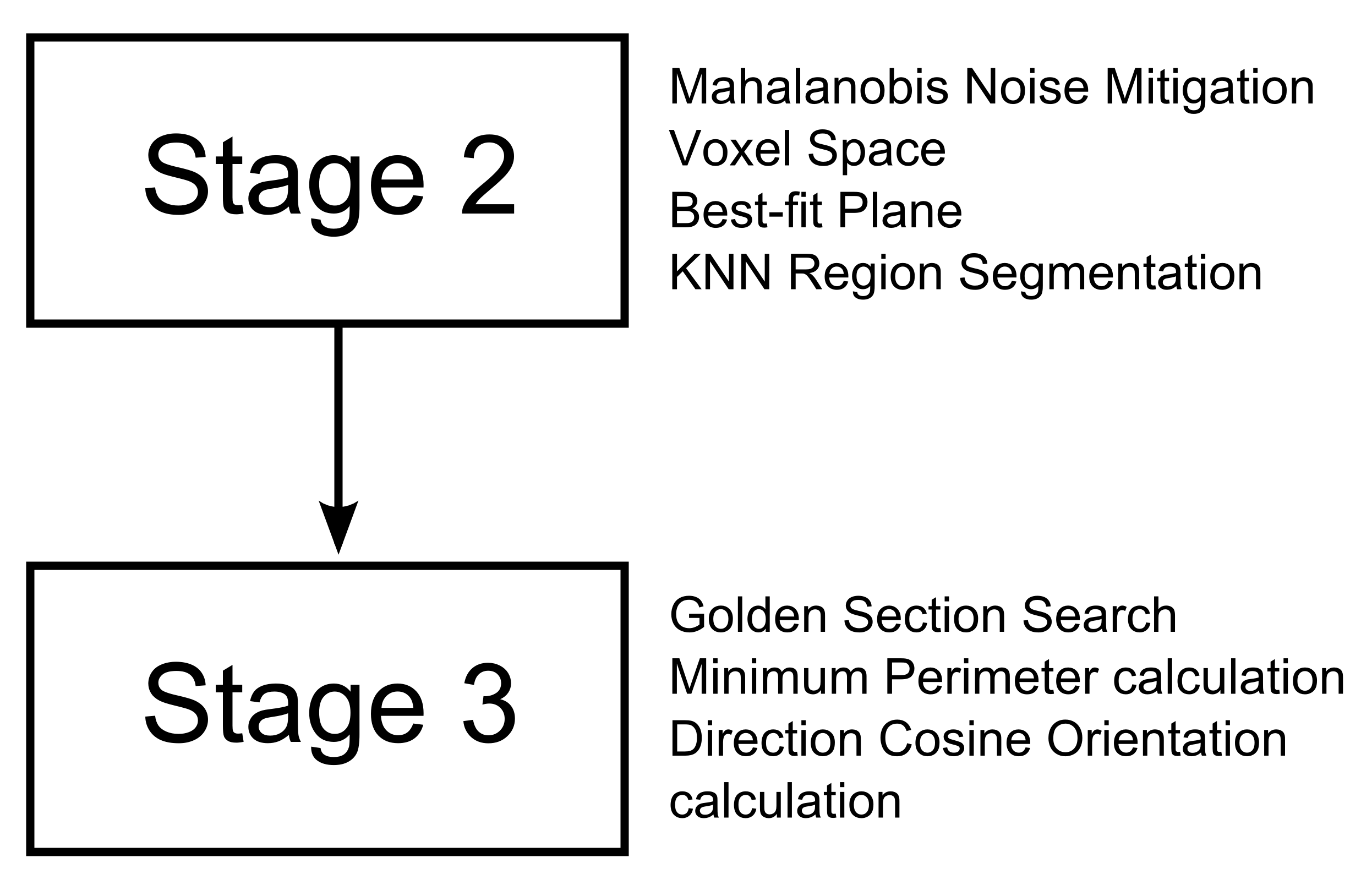}
	\caption[Software Toolchain Workflow]{Outline of functions handled explicitly by the GeoStructure.m pipeline.}
	\label{fig:geostructure flow}
\end{figure}

\subsection{Structure from Motion and Integration with Geology}
SfM extracts structure from sequences of images and is a subset of the Multi-View Geometry (MVG) paradigm, similar to 3D reconstruction from stereo vision. Westoby et al.~\cite{Westoby2012} showed the effectiveness of SfM Digital Elevation Models (DEMs) across a variety of topographic types, demonstrating that SfM DEMs are accurate to the decimetre. This is echoed work by other researchers~\cite{Hartley2003,Harwin2012,James2012}.

\subsection{SfM Data Acquisition: Exploring Camera Paths}
Various studies in the field have considered what configuration of camera views are optimal and how factors impacting the views, influence the final reconstruction. DallAsta et al.~\cite{DallAsta2015} used a Monte Carlo based protocol to test the differences between sampling using a straight line through a scene, or a circular path with all views converging on the same area. They also considered other factors, such as the use of ground control points, autofocus and camera calibration. James et al.~\cite{James2014} suggest that relying on automatic camera calibration can lead to systematic residual errors in the final reconstruction but also develop methodologies to mitigate these effects. Of particular interest to this work are the effect of camera path, and factors generating radial distortion effects~\cite{DallAsta2015, Bemis2014, James2012, Vasuki2014}.

\subsubsection{Mitigating Radial Distortion}
James et al.~\cite{James2014} specifically reference that radial distortion occurs as a result of a combination of near-parallel imaging directions and poor correction of radial lens distortion. This is referred to as "doming" because it generates a parabolic-type distortion in the reconstructed surfaces. To mitigate this, the inclusion of differing angles of the object/scene as well as oblique images are shown to be effective~\cite{DallAsta2015, Bemis2014, James2012, Vasuki2014}. This highlights the importance of "flight plan" when acquiring data for a reconstructive process such as ours, and was implemented here.

\subsection{Region Growing}
The work of Rabbani~\cite{Rabbani2006} into point cloud processing and segmentation, demonstrated a framework for a fully automatic solution for this process. This work utilised a combination of a $k$ Nearest Neighbour Region Growing algorithm, combined with a surface normal calculation incorporating a smoothness constraint.

\subsection{Filtering Noise and Erroneous Points}

Noise is a major concern when dealing with software generated point clouds. Erroneous points are an artifact of the mathematical processes used to generate the reconstruction from views. There are a number of ways to mitigate such artifacts. Xi~\cite{Xi2009} used a non-parametric method incorporating an anisotropic kernel density estimation for outlier removal, along with a hill climbing line search for accurate approximation of the real surface boundary. In contrast, Brophy~\cite{Brophy} uses a Mahalanobis distance algorithm to remove outlier and noise points from a preprocessed point cloud while maintaining high fidelity with the original structure.
\begin{figure}[h]
	\centering
	\includegraphics[width=0.45\textwidth]{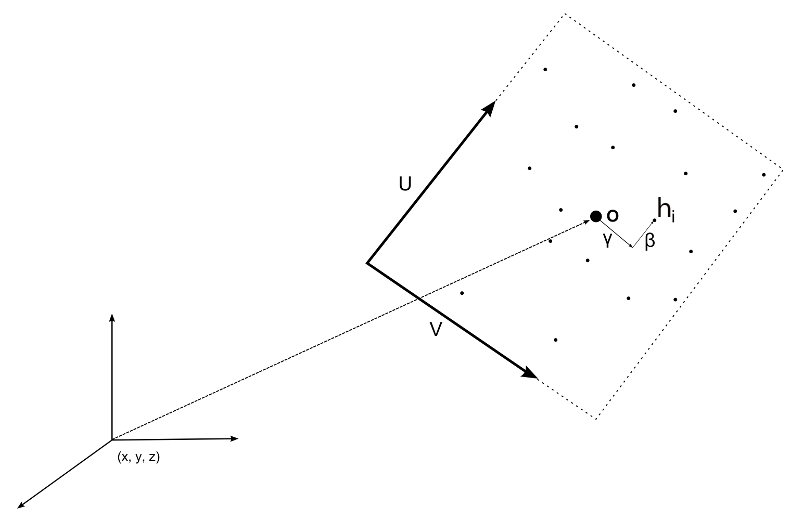}
	\caption[Traversing the Region Plane]{The factors $\beta$ and $\gamma$ describe the magnitude of direction of the point, $h_{i}$, from the origin, $o$, in the $(u,v)$ coordinate system or \textit{region space}.}
	\label{fig:beta gamma plane}
\end{figure}
\section{GeoStructure Software Toolchain Overview}

The GeoStructure Software Toolchain encompasses the whole process from data acquisition to Planar Orientation extraction. The Software Toolchain has three stages: Data Preprocessing, Data Processing and Data Output/Orientation Measurement. The approach is summarised below:

\begin{itemize}
	\item Utilise a high sample rate strategy (video)
	\item Convert video to image frames (25fps)
	\item Process images to produce sparse point cloud
	\item Process sparse point cloud to produce dense point cloud
	\item Process dense point cloud to produce highly defined surface (as ~\cite{Gigli2011})
	\item Develop algorithm extract Planar Orientation measurements of major surfaces
	\begin{itemize}
		\item Process noise and outlier points
		\item Apply a voxelisation method on the point cloud data
		\item Define large scale regions for measurement
		\item Find optimum perimeter for each region
	\end{itemize}
\end{itemize}

The Toolchain flow is illustrated by Fig~\ref{fig:toolchain dataflow}. The following section briefly describes each component.

\subsection{Stage 0: Data Acquisition}
We contend that we can increase the number of images in the sample set and produce a reconstruction of fidelity high enough, to measure the reconstructed surfaces to scientifically valid standard directly, rather than having to utilise any kind of transformative inference (such as the Hough Transform) to obtain 3D measurements from a 2D source~\cite{Brown2002,Guerrero2018,Kemeny2003,Kissi2016,Lowe1999,Lowe2004,Post2001, Rabbani2006,Vasuki2014,Vosselman2004,Wang2009}. 

Utilising video as the capture method~\cite{McLeod2013}, allows for reasonably simple acquisition strategy, while ensuring a dense spread of views and angles. Another reason we obtain a dense spread of views is because previous studies have shown that good SfM reconstructions generally come from images overlapping by approximately 50-60\%, with high value placed on oblique and converging images~\cite{Bemis2014,James2012,James2014}. In this case, splitting the video frames, as well as the camera path, ensures the image overlap and obliquity are present in the image set. This path is shown in Fig~\ref{fig:example sparse flight plan}, which illustrates how the bundle adjustment process also reconstructs the positions of the views used in the reconstruction, with the high density of view samples shown.

\subsection{Stage 1: Preprocessing}
An image-based SfM software (VisualSfM~\cite{Wu2011,Wu2013}) was used and a small program (contained within GeoStructure) was used to parse out frames at the standard 25 fps to jpg format.

\subsection{Stage 2: Initial Point Cloud generation and Densification}
The next data processing stage is the SfM treatment using the VSfM software package~\cite{Wu2011,Wu2013}. This stage of the Toolchain begins with the preprocessed image set, and ends with the sparse point reconstruction of the scene. This is then densified via a VSfM plugin: PMVS.  Despite this process selecting out some extraneous points, this part of the process still generates noise that must be mitigated later in the process. 

\subsubsection{Stage 2: Surface Tessellation using Poisson Surface Reconstruction}
The addition of noise during the previous stages leads to a compounding of noise artifacts, that result in erroneous tessellations and false surfaces in the final rendering.

The Poisson vector field tessellation is used generate a solid surface from the dense reconstruction. The PSR method has some unique characteristics useful in this case, such as the property of treating the whole point cloud simultaneously~\cite{Kazhdan2006,Kazhdan2013a,Calakli2011}. This leads to smoother surfacing results~\cite{Kazhdan2013a}, which is of particular value considering the intrinsic error generation of this point cloud method.

\begin{figure*}[h]
	\centering
	\begin{subfigure}[b]{0.24\textwidth}
		\includegraphics[width=1\textwidth]{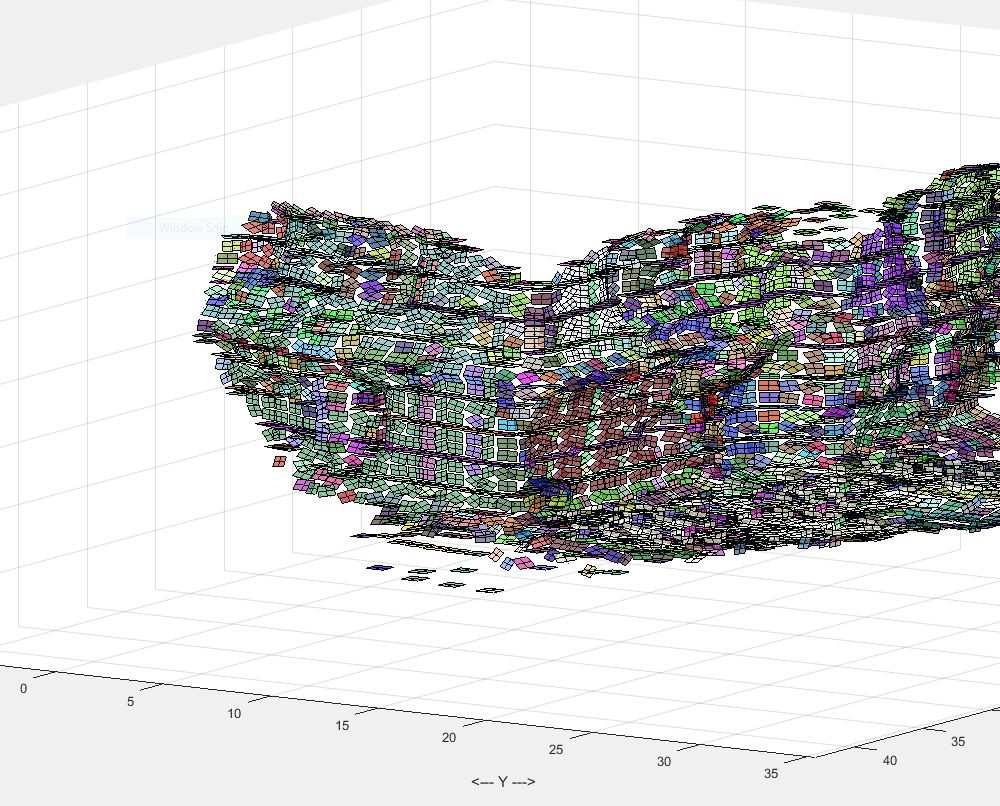}
		\caption{Area corresponding the \textit{Region 1} prior to initialisation of Golden Section Search.}
		\label{fig:phi planes start}
	\end{subfigure}
	~ 
	\begin{subfigure}[b]{0.24\textwidth}
		\includegraphics[width=1\textwidth]{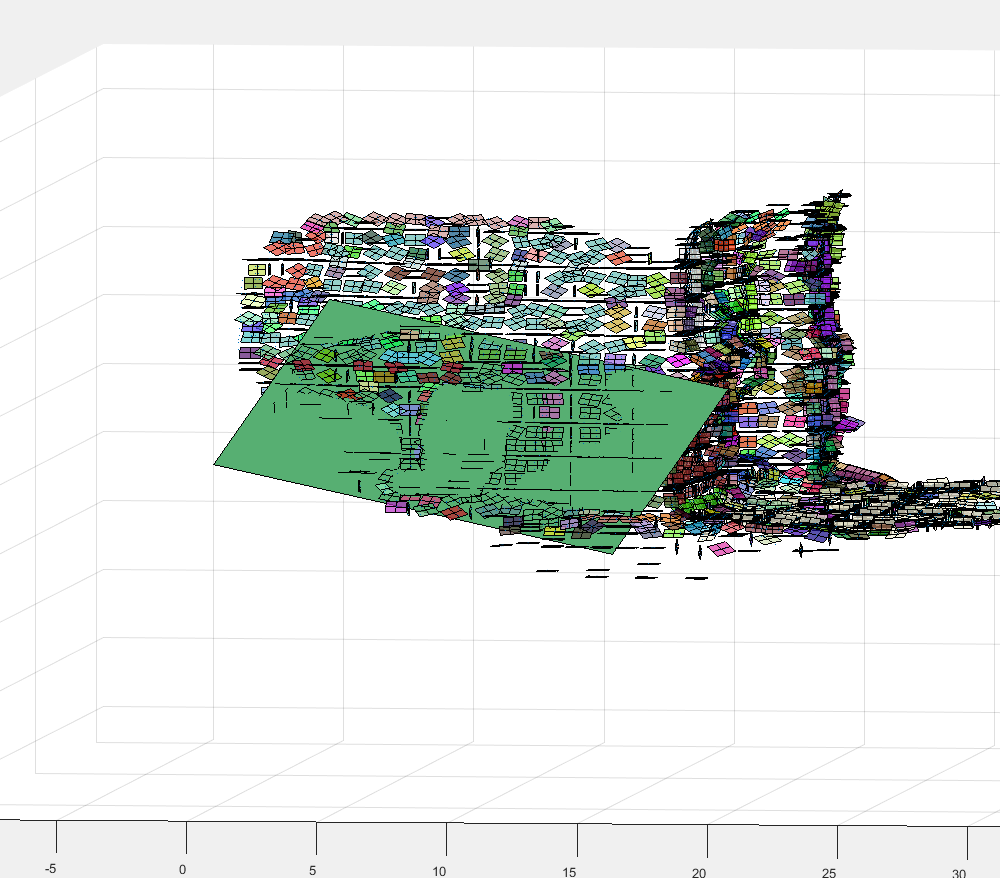}
		\caption{First candidate plane geometry.}
		\label{fig:phi planes 1}
	\end{subfigure}
	~ 
	\begin{subfigure}[b]{0.24\textwidth}
		\includegraphics[width=1\textwidth]{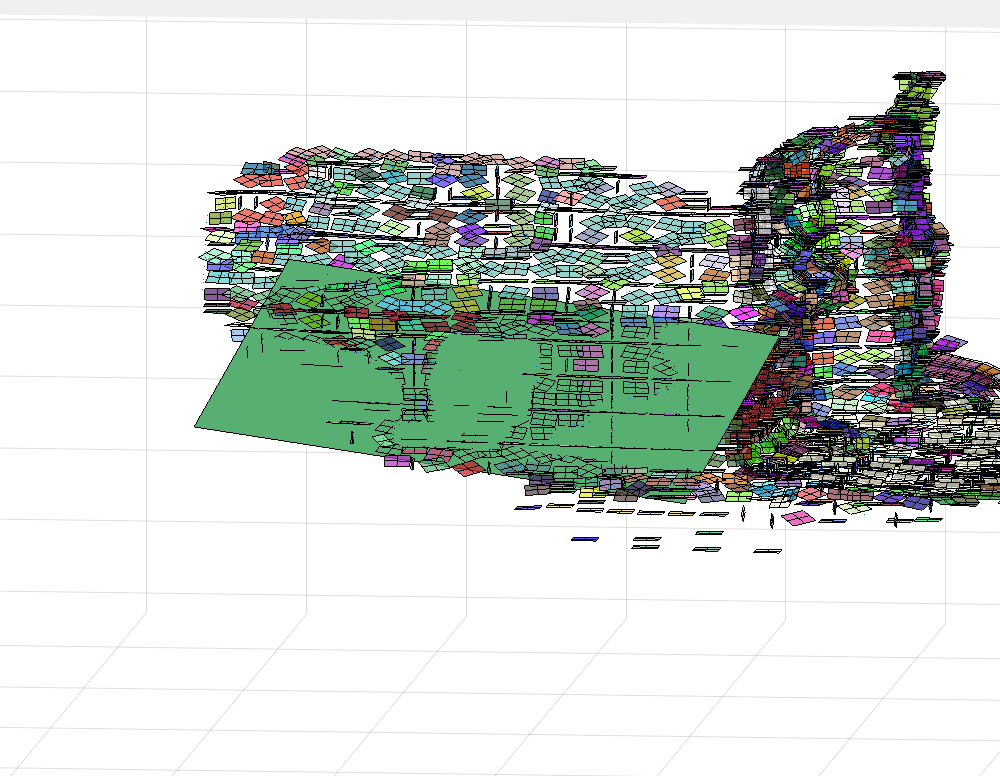}
		\caption{Sixth candidate plane geometry.}
		\label{fig:phi planes 2}
	\end{subfigure}
	\begin{subfigure}[b]{0.24\textwidth}
		\includegraphics[width=0.9\textwidth]{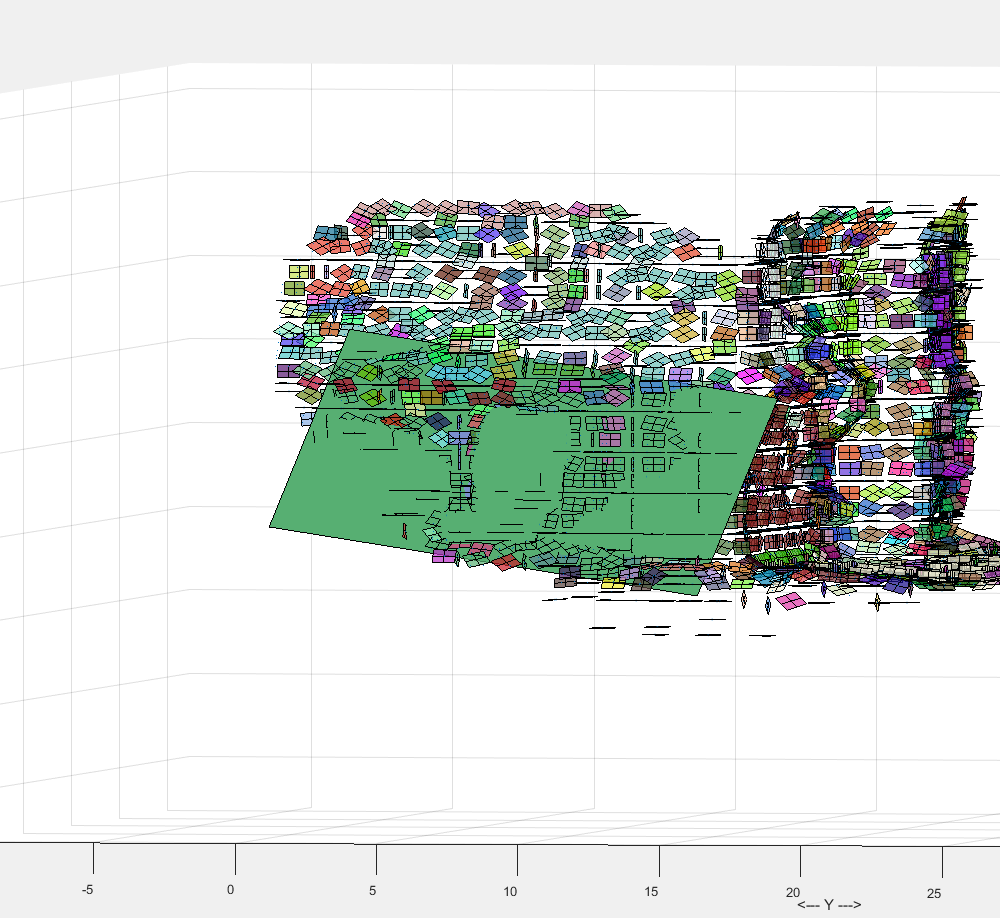}
		\caption{Ninth candidate plane geometry.}
		\label{fig:phi planes 3}
	\end{subfigure}
	\begin{subfigure}[b]{0.24\textwidth}
		\includegraphics[width=1\textwidth]{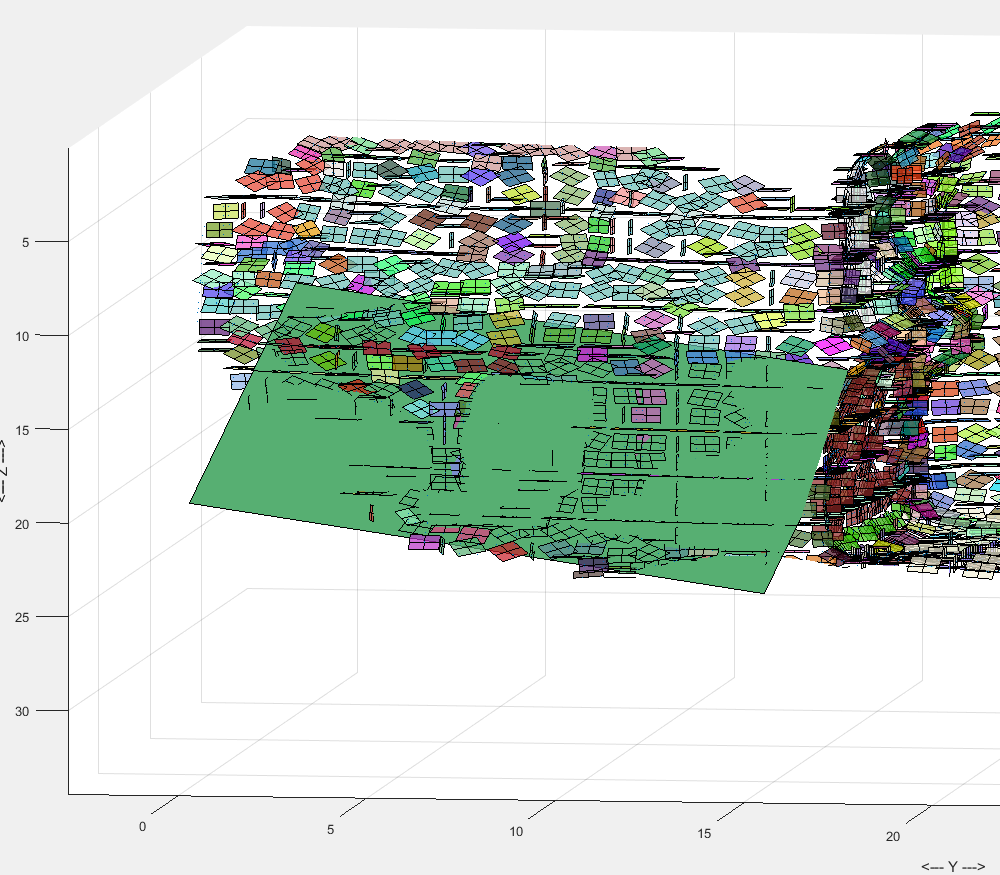}
		\caption{The final orientation and size of the plane for \textit{Region 1}.}
		\label{fig:phi planes end}
	\end{subfigure}
	\begin{subfigure}[b]{0.24\textwidth}
		\includegraphics[width=1\textwidth]{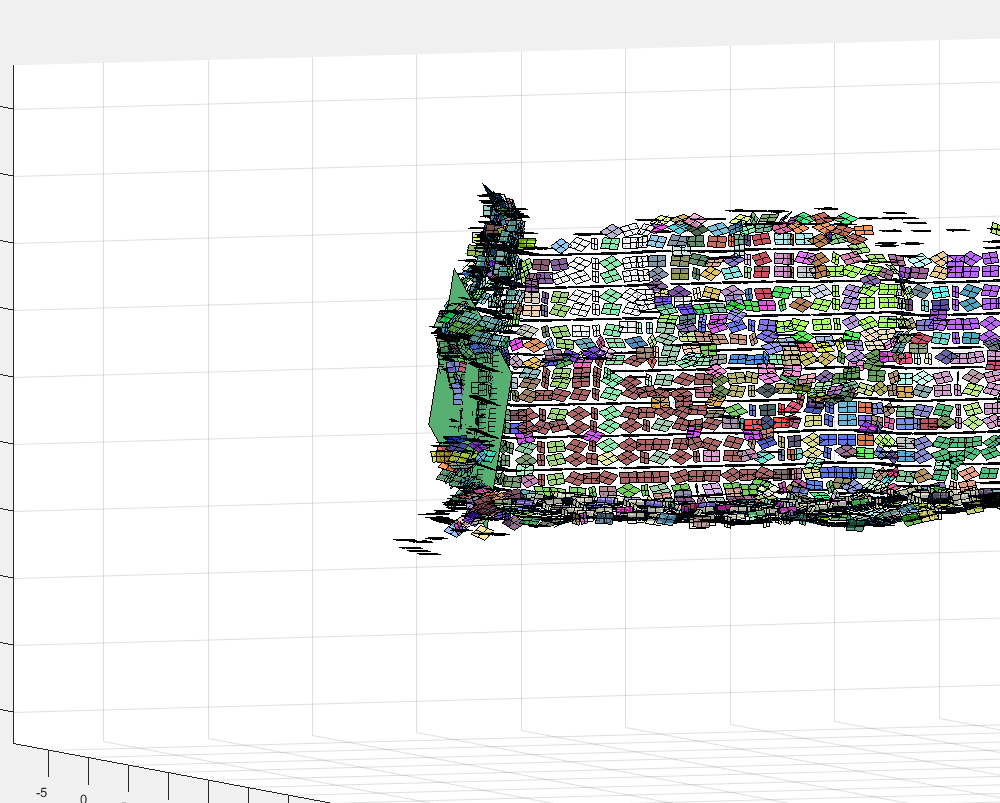}
		\caption{First candidate plane geometry.}
		\label{fig:phi planes 1e}
	\end{subfigure}
	\begin{subfigure}[b]{0.24\textwidth}
		\includegraphics[width=1\textwidth]{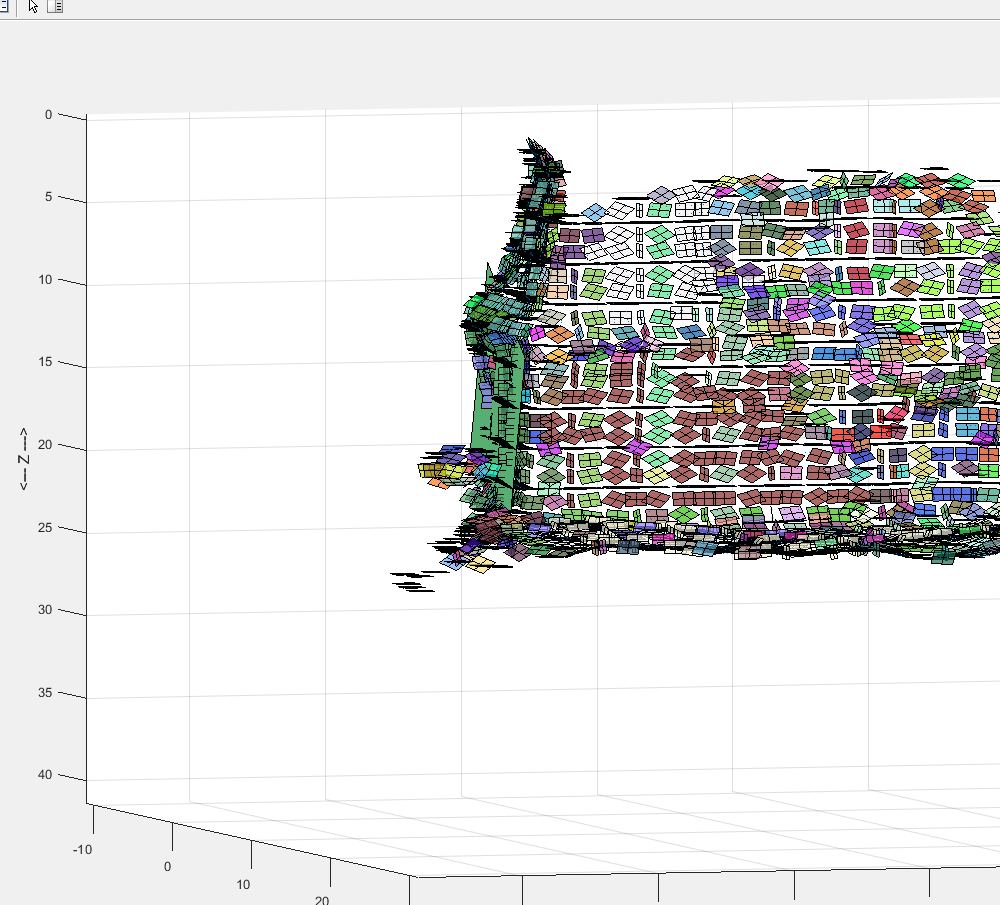}
		\caption{Sixth candidate plane geometry.}
		\label{fig:phi planes 2e}
	\end{subfigure}
	\begin{subfigure}[b]{0.24\textwidth}
		\includegraphics[width=1\textwidth]{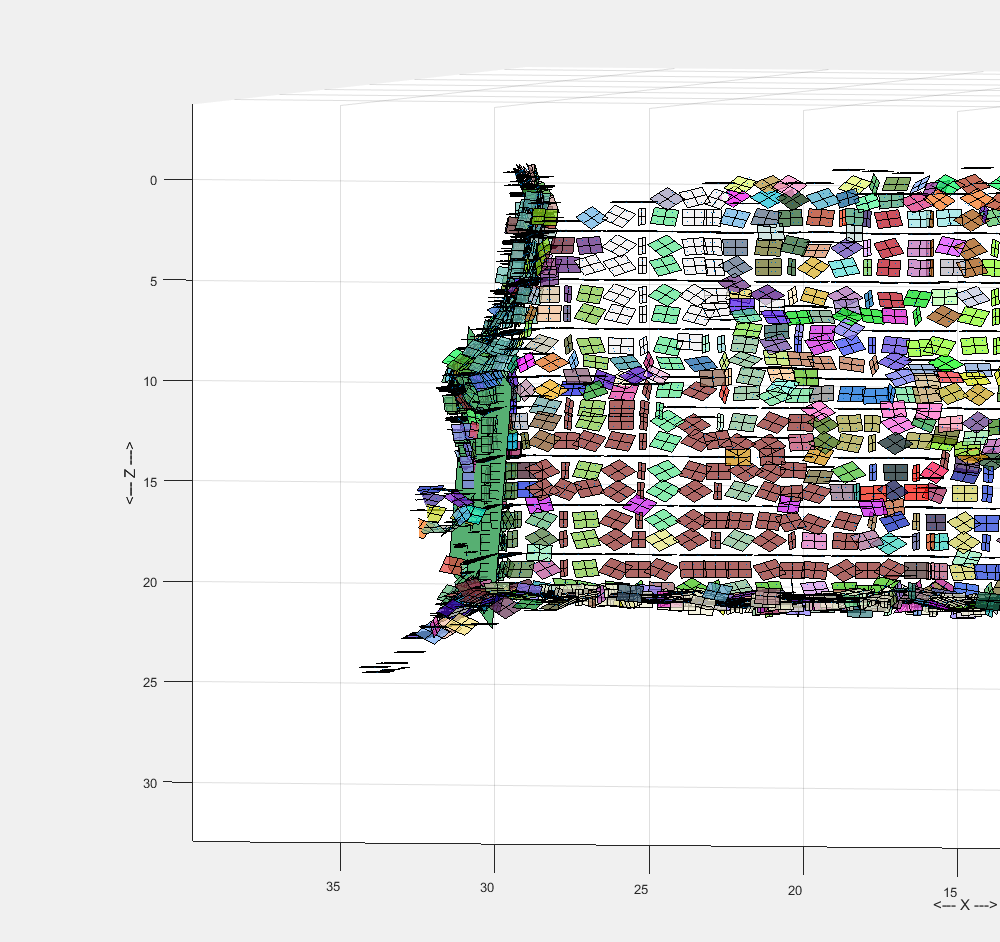}
		\caption{Ninth candidate plane geometry.}
		\label{fig:phi planes 3e}
	\end{subfigure}
	\caption{The above figure shows the process of Region Plane definition from the mathematically derived boundless \textit{Ideal Plane} via the use of the $\phi$ Golden Section Search and Rodrigues formula Rotation Matrix. For this particular cycle, the total number of candidate plane geometries iterated through numbered 14, with the 14th being the final \textit{Region Plane}  corresponding to the Region 1 ground-truth (See Fig.~\ref{fig:gt six surfaces}). Top images show rotational/translational/stretch variance cycled through by the $\phi$ search, while the bottom images show the some of the normal variation. Left to right, top to bottom: \ref{fig:phi planes start}/\ref{fig:phi planes end}: No region candidate/bounded Region Plane 1, \ref{fig:phi planes 1}/\ref{fig:phi planes 1e}: Candidate geometry 1, \ref{fig:phi planes 2}/\ref{fig:phi planes 2e}: Candidate geometry 6, \ref{fig:phi planes 3}/\ref{fig:phi planes 3e}: Candidate geometry 9. }
	\label{fig:phi planes candidate}
\end{figure*}

\section{Stage 2: GeoStructure Pipeline}
The GeoStructure pipeline design is the result of analyses of previous work of researchers in geology, machine learning/algorithm development, image processing and computer vision~\cite{Gigli2011,Rabbani2006,Furukawa2010a,Wu2013,Snavely2006,Vasuki2014,Bemis2014}, with the  aim of extracting accurate measurements directly from a software generated point cloud itself, with minimum input/correction from the user.

\begin{figure}[h]
	\centering
	\begin{subfigure}[b]{0.2\textwidth}
		\includegraphics[width=\textwidth, height=35mm]{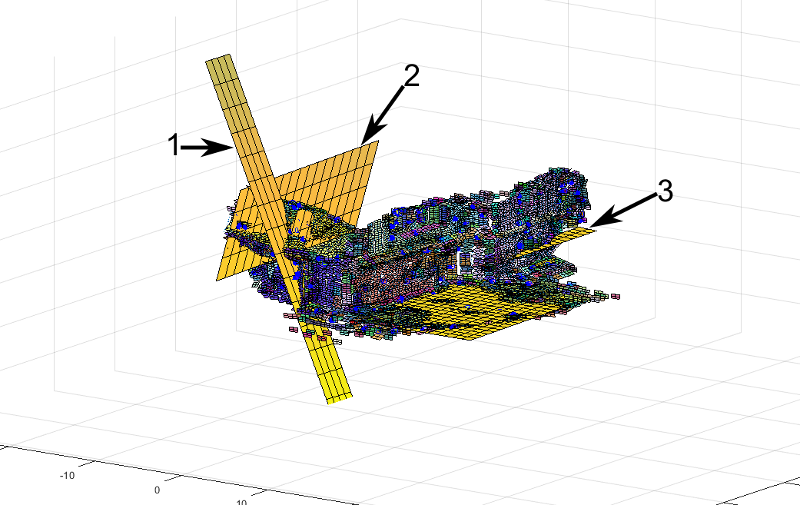}
		\caption{(x, y, z)}
		\label{fig:ortho planes}
	\end{subfigure}
	~ 
	\begin{subfigure}[b]{0.2\textwidth}
		\includegraphics[width=\textwidth, height=35mm]{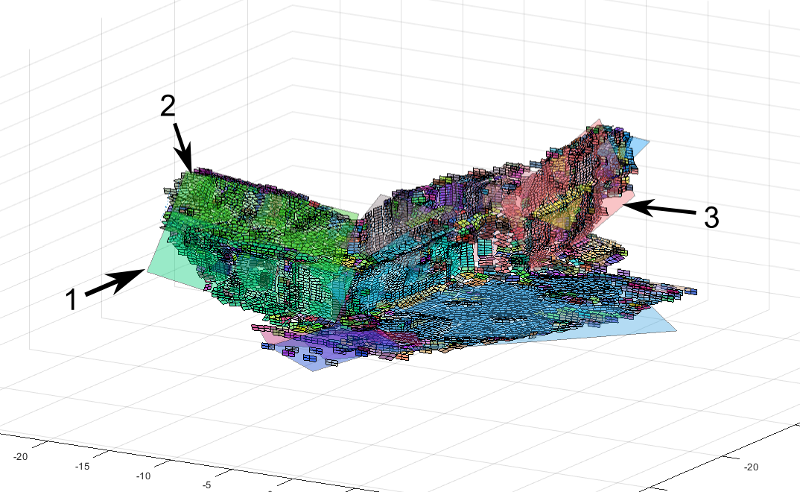}
		\caption{(u, v, n)}
		\label{fig:phi planes}
	\end{subfigure}
	\caption[$\varphi$ Golden Section Method]{The results of using standard World reference for drawing planes (a) and using the $\varphi$ Golden Section Minimum Perimeter Search (b). The same regions are labelled between both. (b) is visibly more accurate, with colours of the region planes matching the underlying voxel planes that make up the region}
	\label{fig:phi section planes}
\end{figure}

Below is a summary of the function of the GeoStructure at each processing stage:
\begin{enumerate}
	\item Read point cloud file 
	\item Create point cloud volume [point space] of uniform dimensions
	\item Search the volume, one voxel at a time
	\item Evaluate each voxel for the presence of points
	\item If points are present, attempt to fit a best-fit plane to the volume
	\item Segment to the point cloud by grouping planes into macroscale surface regions
	\item Aggregate planes of same region into a singular plane approximating the overall best fitting orientation
	\item Extract orientation measurements from regions. 
\end{enumerate}

The GeoStructure pipeline flow is illustrated by Fig~\ref{fig:geostructure flow}. The following section briefly describes each component.

\begin{figure*}
	\centering
	\includegraphics[width=0.95\textwidth]{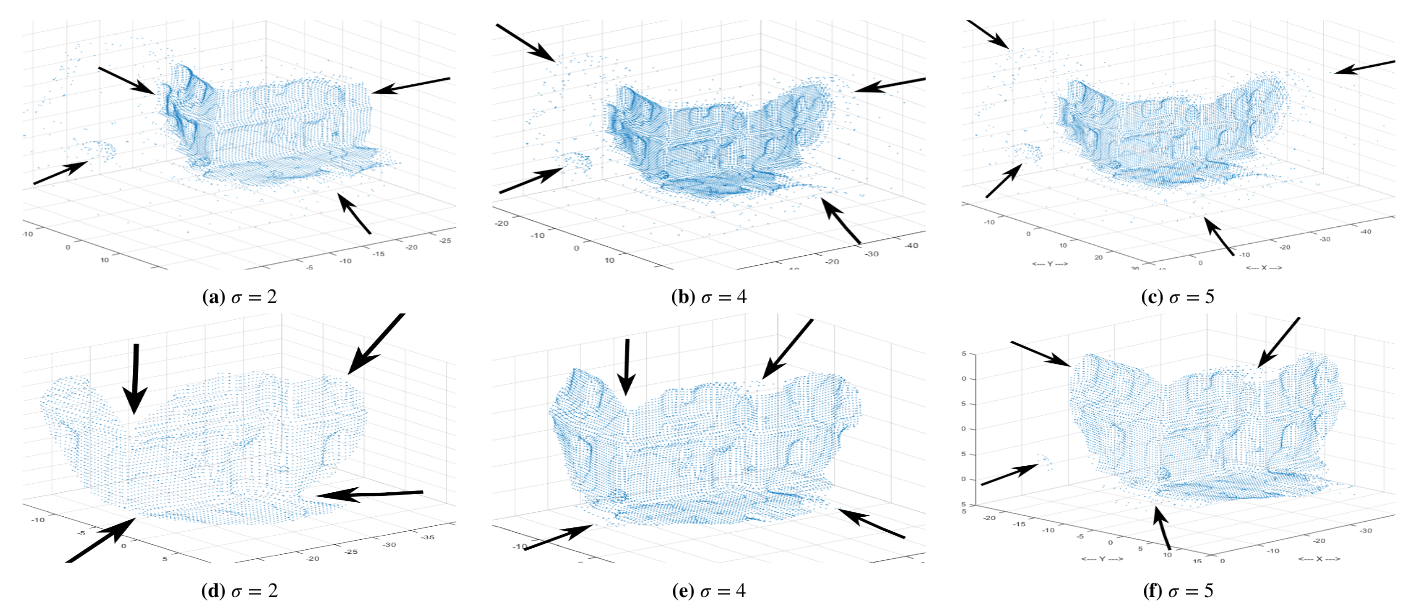}
	\caption{The above reconstructions compare the effect of a standard Euclidian distance method (a, b, c) against the vanilla Mahalanobis method (d, e, f). The Euclidean method shows large numbers of noise clusters at a distance of 4 \& 5 standard deviations ($\sigma$) from the mean. At $\sigma$ = 2, many noise clusters still occur, overall structural point density visibly down and some legitimate structure no longer present. In contrast, the Mahalanobis distance demonstrates significantly better behaviour at all $\sigma$ values. Noise clusters are greatly reduced at $\sigma$ = 4 \& 5, although some minor structural erosion has occurred. At $\sigma$ = 2 there is more significant erosion but all noise has been filtered.
	}
	\label{fig:noise}
\end{figure*}

\subsection{Mitigating Anisotropic Noise}
When reduced back to a point cloud, artifacts generated during the PSR phase will have a generally anisotropic-type sparse noise profile. Brophy~\cite{Brophy} demonstrated the effectiveness of a Mahalanobis distance metric for removing anisotropic-type noise from point clouds.

The Mahalanobis distance operates in the following way:
\begin{equation}
d_{\Sigma}(x, x_{i}) = ((x - x_{i}^{\intercal})H^{-1}(x - x_{i}))^{1/2},
\end{equation}
where the covariance matrix 
\begin{equation}
H = DD^{\intercal},
\end{equation}
is constructed using the following:
\begin{equation}
D = (x_{1} - x, x_2 - x, \dotsc, x_{n} - x).
\end{equation}
where $x$ is the total points in the point cloud. It provides a measure of the distance between a point in the set and local distributions of the set, along each principle component axis. It has the property of unitless scale invariance, and respects dataset correlations~\cite{Brophy,Xi2009,Snavely2006}. This has the effect of delineating between points that are outliers and points that are sparsely distributed but still part of the targets physical structure. In contrast, a standard Least Squares-type distance noise filters cannot discriminate, and so will remove points that are desired. Illustrated in Fig.~\ref{fig:noise}

\subsection{Stage 2: The Voxel-fit Process}
In a similar fashion to many works~\cite{Gigli2011, Gomez2014, Turner2013, Vosselman2004, McLeod2013,Furukawa2010a,Calakli2011,Kazhdan2006,Kazhdan2013a,Mah2011,Brophy,Seitz2006}, we utilise a voxelisation strategy to search the point space. A unitless factoring value ($\zeta$) is used to indicate the desired voxel resolution. $\zeta$ is a linear parameter that leads to a cubic expansion of the in the number of voxels.

\subsubsection{Applying the Voxel-fit}
In each voxel, the best-fit plane is calculated (Fig.~\ref{fig:geostructure full}). All the points in the volume are then minimised with respect to the newly defined plane point, producing a modified point sample. The principle directions are then computed through a standard \textit{eigenvector decomposition} (utilising matrix transpose multiplication), from which the point normal and orthonormal bases are extracted. This process is repeated for all voxels.

\subsection{Stage 2: Segmentation through $k$NN Region Growing}

The result of the voxel-fit stage is a series of planes that approximate the surface of the point cloud (Fig.~\ref{fig:geostructure full}). The number of planes make PO measurement unfeasible, so the next step is to use a $k$ Nearest Neighbour Region Growing algorithm~\cite{Rabbani2006}, to aggregate these planes~\cite{Gigli2011} to macroscale \textit{Regions}.

This methodology utilises a threshold, $\theta$, that describes the angle where deviation from the previous plane is considered a new surface, and, a surface offset residual threshold, $\psi$, that represents co-planiarity between planes with similar values for $\theta$. Together, these two variables allow the algorithm to be tuned to delineate differing surfaces for the region growing process (Fig.~\ref{fig:geostruct pc rg}).

\subsection{Stage 3: Building the Region Plane}
The ultimate aim of the $k$NN algorithm is to identify and mark all the voxel planes that share orientation, within a certain tolerance (denoted by $\theta$ and $\psi$). Once groups of like planes are identified, a process of "Region Growing" is initiated that results in the fitting of a \textit{Region Plane} to the voxel regions. 

A simple aggregation of the voxel planes, leads to aggregation of all the residual errors forming highly erroneous Region Planes (Fig.~\ref{fig:ortho planes}), so the process of forming accurate Region Plane is more involved. The complete process of building these planes requires: 

\begin{itemize}
	\item Calculation of the plane centroid/normal (which defines the "Ideal Plane")
	\item Calculation of the minimum perimeter, utilising a rotation matrix, with values defined by the Golden Section Search
\end{itemize}

In order to generate region planes that more accurately represent the configuration of underlying points and avoid the effects of residual errors, we move the calculation from the World coordinate reference frame ($x, y, z$) to a local coordinate reference or \textit{region space} ($(x, y, z) \rightarrow (u, v, n)$) shown in Fig.~\ref{fig:beta gamma plane}).

We define an Ideal Plane to be a boundless plane geometry with centroid at the centre of the region proposed by like voxel planes and [normal] orientation averaged from those voxel planes. Within this structure, there are an infinite number of possible oriented bounded geometries that can be defined as the final Region Plane.

We then make two assumptions: for simplicity, we assume the geometry of the region plane to always be rectangular and enforce this geometry. Secondly, we assume there is an optimum rotation of the Ideal plane in this $u, v$ coordinate system that minimises the perimeter, at which point the Ideal plane can be bounded to form the final Region Plane. 

A Rodrigues skew symmetric rotation matrix, orientated using a Golden Section Search algorithm, is used to find this lowest energy, optimum configuration for each plane (See Fig.~\ref{fig:phi planes candidate}). The result of an attempt to form Region Planes using just World coordinates are shown against our method in Fig.~\ref{fig:phi section planes}).

\begin{figure}[htb]
	\centering
	\includegraphics[scale=.30]{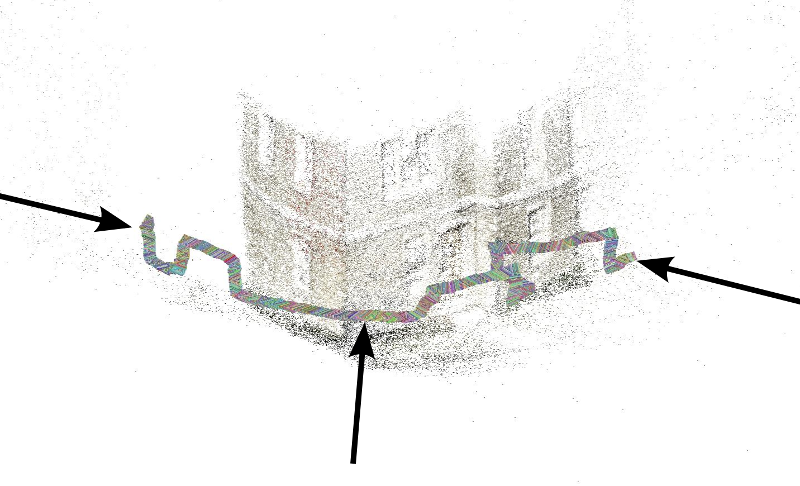}
	\caption[Camera Flight Plan Reconstruction]{The bundle adjustment process reconstructs not only the subject but also the positions of the views used in the reconstruction. The flight path of the camera used is visible and highlighted in this example. The high density of view samples is also apparent, being as it appears as one continuous line.}
	\label{fig:example sparse flight plan}
\end{figure}

\subsection{Stage 3: Obtaining Planar Orientations}
Once region planes, with centroid points and normals have been derived, the calculations of PO are relatively simple to apply. A \textit{direction cosine} methodology utilising elements in design and implementation from both Gigli et al.~\cite{Gigli2011} and Feng et al.~\cite{Feng2001} is used here. This method allows the normal vector of a surface to be used to calculate PO measurements. The derivation of the normal vector is instead done using the standard eigenvector matrix transpose multiplication method, as mentioned previously.

\section{Experimental Framework}
The primary purpose of this experiment set is to determine accuracy of the PO measurements obtained from the reconstruction as compared to pre-collected ground-truth measurements. For interpretation, PO are refrerred to using $P$, with Strike, Dip Angle and Dip Direction being delineated by $P_{\varsigma}$, $P_{\alpha}$ and $P_{\delta}$, respectively.

\begin{table}[H]
	\centering
	\caption{Table of Variable Factor values}\label{tab__varfac_values}
	\begin{tabular}{cccc }
		\toprule
		Var. Factor & Start & End & Step\\
		\midrule
		$\zeta$ & 0.01 & 0.07 & 0.006 \\
		$\theta$ & 0 & 30 & 3 \\
		$\psi$ & 0.01 & 0.6 & 0.06 \\
		$k$ & 1 & 20 & 2 \\
		\bottomrule
	\end{tabular}
\end{table}

For a simple proof of concept for this system, The Hume Cronyn Memorial Observatory building on the Western University campus was selected as its [$P$] dimensions are easily ground-truthed. Its overall size and the size of its features are comparable to structure a geologist would find in the field.

Data was collected maintaining a [roughly] parallel distance of 3-4 metres (see Fig.~\ref{fig:example sparse flight plan}), while remaining cognisant of utilising strategies mentioned previously to reduce doming and other artifacts~\cite{Bemis2014,Vasuki2014,Snavely2006,Westoby2012,DallAsta2015,James2014,James2012}.

The dataset consisted of 1~min 25~secs of video in .avi format, taken with the Canon Ixus 8.0 Mpixel in video mode at 640 $\times$ 480 pixel resolution. This decomposed into 2573 individual frames, at approximately 25-30 fps, which was then fed into the VSfM package as the sample set.

\begin{figure}[h]
	\centering
	\begin{subfigure}[b]{0.25\textwidth}
		\includegraphics[width=1\textwidth,height=25mm]{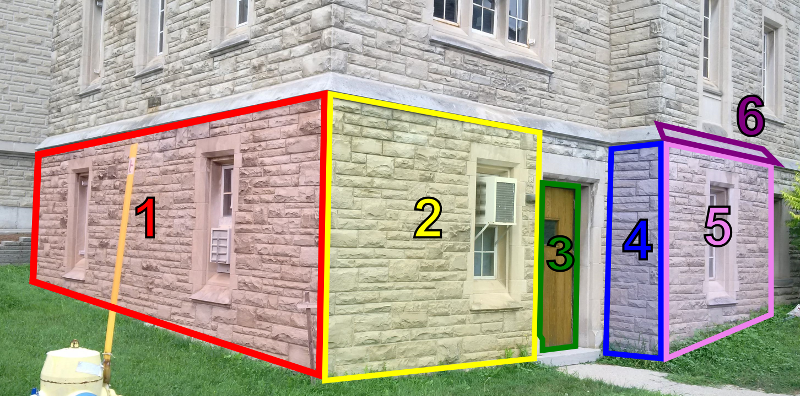}
		\caption{}
		\label{fig:gt six surfaces}
	\end{subfigure}
	\begin{subfigure}[b]{0.15\textwidth}
		\includegraphics[width=1\textwidth,height=25mm]{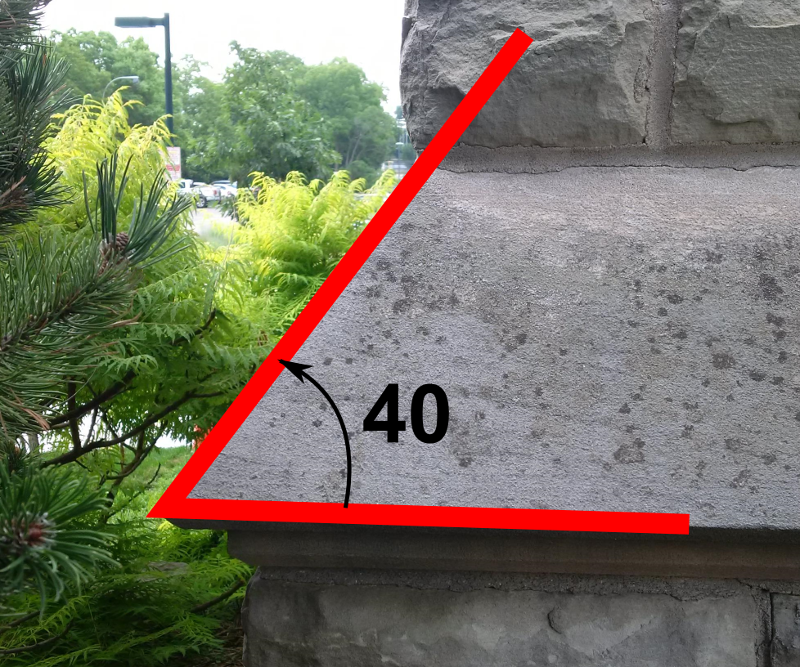}
		\caption{}
		\label{fig:gt surface 6 angle}
	\end{subfigure}
	\caption[GeoStructure.m Processing]{The 6 surfaces selected for ground truth measurement are highlighted in this image. The Planar Orientations are detailed in Table~\ref{tab__gt_values}. A close-up image of the inclination of surface 6. The upward normal from the horizontal is the dip angle recorded. In this case, the angle is 40$^{\circ}$.}
	\label{fig:gt six surfaces2}
\end{figure}

\section{Results Framework} \label{results_interpretaion}

The efficacy of the pipeline is analysed through two methods. The first is recording processing time, with particular attention paid to what fraction is the result of region growing. Region growing is the single most complex part of the process and so is reasoned that completion time of this part will be proportionally representative of the completion time of GeoStructure as a whole. The second method will be the effect of \textit{Variable Factor} variation on each of the six selected region planes, using a \textit{Quality Metric}.

\begin{table}[h]
	\centering
	\caption{Table of Ground-truth Values}\label{tab__gt_values}
	\begin{tabular}{cccc}
		\toprule
		Surface & Str($^{\circ}$) & Dip Ang($^{\circ}$) & Dip Dir($^{\circ}$)\\
		\midrule
		1 & 87 & 89(S) & 177 \\
		2 & 04 & 89(E) & 94 \\
		3 & 339 & 87(E) & 69 \\
		4 & 102 & 86(S) & 192 \\
		5 & 350 & 89(E) & 80 \\
		6 & 359 & 40(E) & 89 \\
		\bottomrule
	\end{tabular}
\end{table}

The are four Variable Factors that the GeoStructure pipeline requires:
\begin{itemize}
	\item $\zeta$: The voxel size factor (dictates the size of the voxel planes for each region).
	\item $\theta$: The angle threshold (determines if voxel planes are the same surface [for region growing]).
	\item $\psi$: The residual offset (determines whether voxel planes are the same surface [for region growing]).
	\item $k$: The nearest neighbour value used in the region growing subroutine.
\end{itemize}

Theoretically each of these should have an optimum number and GeoStructure can be configured to find the optimum value for each by varying one and maintaining the others at a constant value. The default values for each are given in Table~\ref{tab__varfac_values}:

\begin{figure*}[h]
	\centering
	\begin{subfigure}[b]{0.24\textwidth}
		\includegraphics[width=1\textwidth]{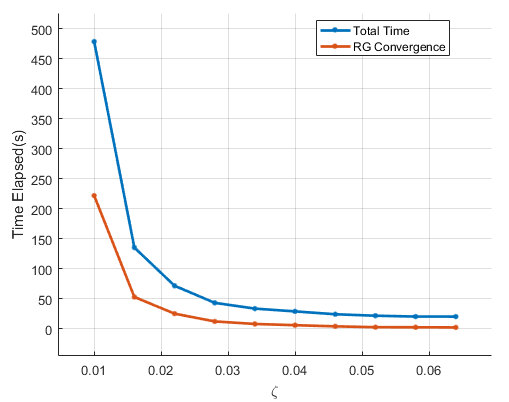}
		\caption{$\zeta$ v time}
		\label{fig:results zeta time}
	\end{subfigure}
	~ 
	\begin{subfigure}[b]{0.24\textwidth}
		\includegraphics[width=1\textwidth]{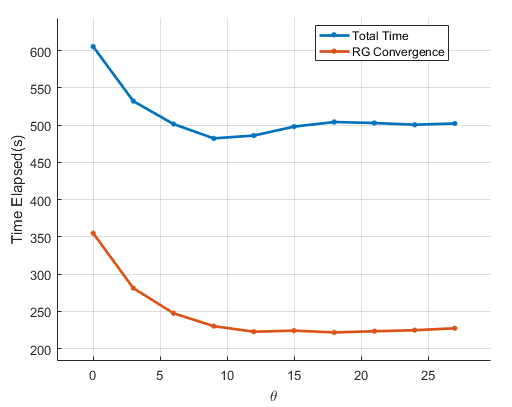}
		\caption{$\theta$ v time}
		\label{fig:results theta time}
	\end{subfigure}
	~ 
	\begin{subfigure}[b]{0.24\textwidth}
		\includegraphics[width=1\textwidth]{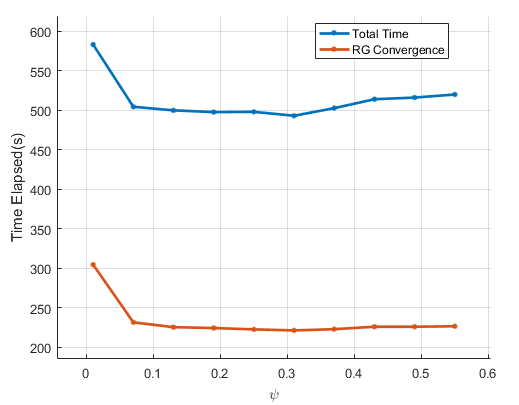}
		\caption{$\psi$ v time}
		\label{fig:results psi time}
	\end{subfigure}
	\begin{subfigure}[b]{0.24\textwidth}
		\includegraphics[width=0.9\textwidth]{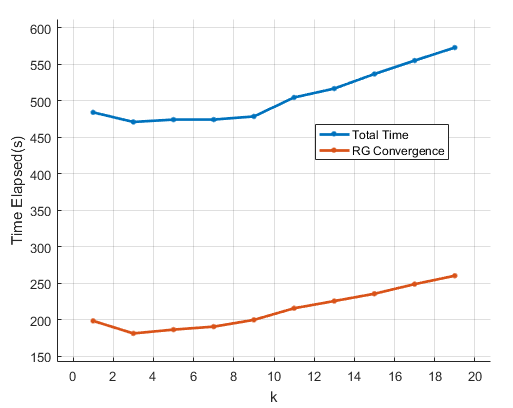}
		\caption{$k$ v time}
		\label{fig:results k time}
	\end{subfigure}
	\begin{subfigure}[b]{0.24\textwidth}
		\includegraphics[width=1\textwidth]{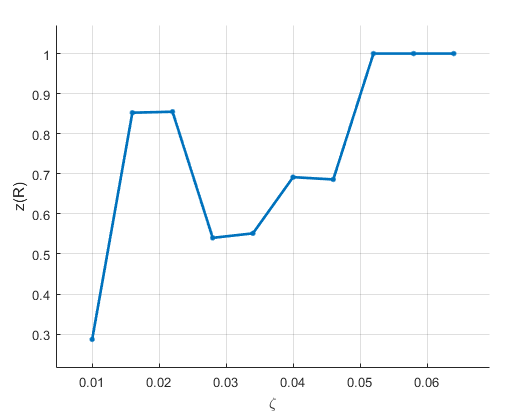}
		\caption{$\zeta$ v $z(R)$}
		\label{fig:results zeta overall z}
	\end{subfigure}
	\begin{subfigure}[b]{0.24\textwidth}
		\includegraphics[width=1\textwidth]{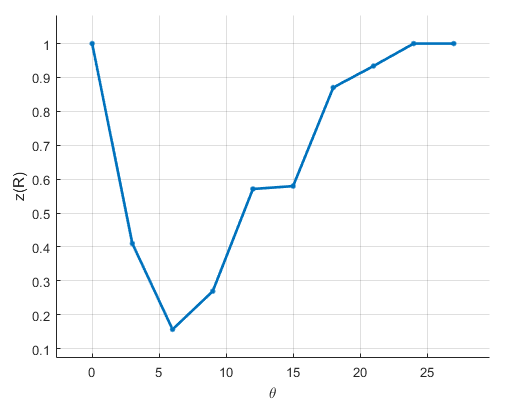}
		\caption{$\theta$ v $z(R)$}
		\label{fig:results theta overall z}
	\end{subfigure}
	\begin{subfigure}[b]{0.24\textwidth}
		\includegraphics[width=1\textwidth]{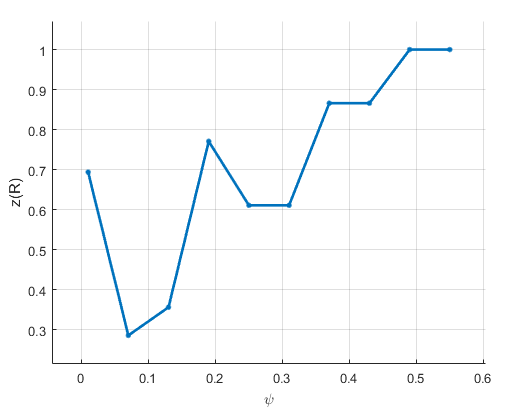}
		\caption{$\psi$ v $z(R)$}
		\label{fig:results psi overall z}
	\end{subfigure}
	\begin{subfigure}[b]{0.24\textwidth}
		\includegraphics[width=1\textwidth]{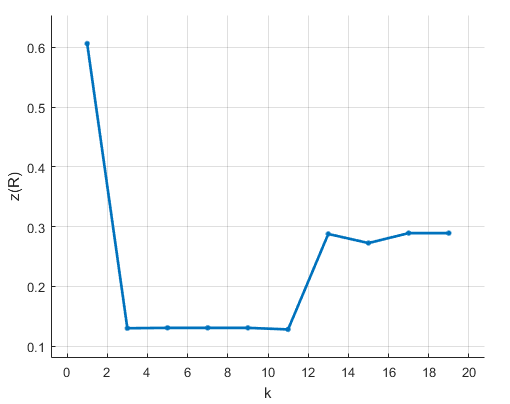}
		\caption{$k$ v $z(R)$}
		\label{fig:results k overall z}
	\end{subfigure}
	\caption{The experimental results of the pipeline output with regards to successive values for each of the Variable Factors (Section~\ref{results_interpretaion}) are shown above. Top: time to completion for both region growing segmentation completion (red) and total pipeline completion (blue). Bottom: overall \textit{Quality Score}, $z(R)$ (as described in Subsection~\ref{quality_score}), with the lower $z$ values indicating the higher measurement accuracy as compared to ground truth.}
	\label{fig:graphs}
\end{figure*}

\subsection{Ground Truth}

Specific surfaces were selected from the structure to be used as $P$ ground-truth. Measurements were obtained using the \textit{Strike and Dip} mobile application, developed by Hunt Mountain Software. Measurements were obtained by laying the device flat, as close to the centre of the surface as possible, with the screen facing the user. The interface can then be used to record the measurement.

\begin{table}[h]
	\centering
	\caption{Table of Ground-truth Surface Normals}\label{tab__gt_norms}
	\begin{tabular}{cc}
		\toprule
		Surface & $\vec{n}$ \\
		\midrule
		1 & -0.9985, -0.0523, 0.0175\\
		2 & -0.0697, -0.9974, 0.0175\\
		3 & 0.3579, -0.9323, 0.0523\\
		4 & -0.9970, 0.0348, 0.0698\\
		5 & 0.1736, -0.9847, 0.0175\\
		6 & 0.0112, -0.6427, 0.7660\\
		\bottomrule
	\end{tabular}
\end{table}

Fig~\ref{fig:gt six surfaces2} shows which surfaces were chosen and Table~\ref{tab__gt_values} and~\ref{tab__gt_norms} show the $P$ measurements input into the pipeline. The long side of the building faces North, which aligns with the $P$ calculation method, which uses the x-axis as the North direction~\cite{Feng2001}, so there is no need for any angular conversion for this dataset.

\subsection{Planar Orientation Quality Score} \label{quality_score}

The ground-truthed and measured region orientations can be compared via a normalisation process, where the angular difference is represented using $z$:
\begin{equation}
z(P_{x}) = \frac{\sqrt{(P_{x\gamma} - P_{x\mu})^{2}} - P_{min}}{P_{max} - P_{min}},  \text{where} 
\begin{cases}
P_{x} \in  \{P_{\varsigma}, P_{\alpha}, P_{\delta}\}, \\
z \in [0, 1] 
\end{cases}
\end{equation}
where $P_{\gamma}$ and $P_{\mu}$ are the \textbf{corresponding} ground-truth and measured quantities and $P_{min}$ and $P_{max}$ are the maximum and minimum possible values in degrees or radians.

We can continue to use this calculation to obtain the  \textit{Quality Score} for a single region surface:

\begin{equation}
z(R_{i}) = \frac{\sqrt{(z(P_{\varsigma}) + z(P_{\alpha}) + z(P_{\delta}))^{2}} - P_{min}}{P_{max} - P_{min}},
\text{where}
\begin{cases}
P_{min} = 0, \\
P_{max} = 3, \\
z \in [0, 1] 
\end{cases}
\end{equation} 

Finally, the same methodology can be generalised to obtain the overall score for the total number regions of that experimental run:
\begin{equation}
z(R) = \frac{\sqrt{\sum_{1}^{n}(z(R_{i})^{2}}) - R_{min}}{R_{max} - R_{min}}, \\
\text{where}
\begin{cases}
R_{min} = 0, \\
R_{max} = n, \\
z \in [0, 1] 
\end{cases}
\end{equation}
This final quantity gives a normalised bounded score for the \textit{run}, which is continuous and directly comparable in all cases. Each \textit{experiment} consists of a series of runs, with a varying Variable Factor, resulting in 4 experiments.

\begin{figure*}
	\centering
	\begin{subfigure}[b]{0.23\textwidth}
		\includegraphics[width=\textwidth]{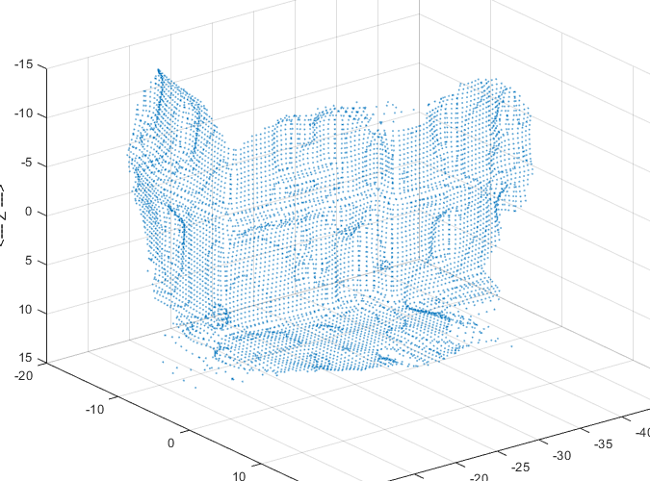}
		\caption{ }
		\label{fig:geostruct pc read}
	\end{subfigure}
	~ 
	\begin{subfigure}[b]{0.23\textwidth}
		\includegraphics[width=\textwidth]{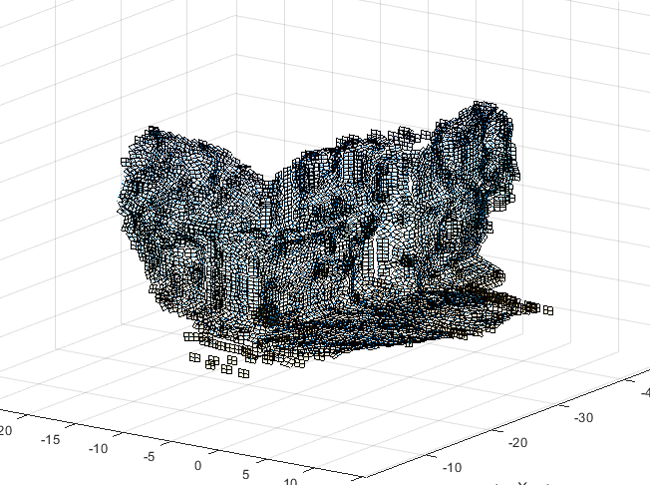}
		\caption{ }
		\label{fig:geostruct pc search}
	\end{subfigure}
	~ 
	\begin{subfigure}[b]{0.23\textwidth}
		\includegraphics[width=\textwidth]{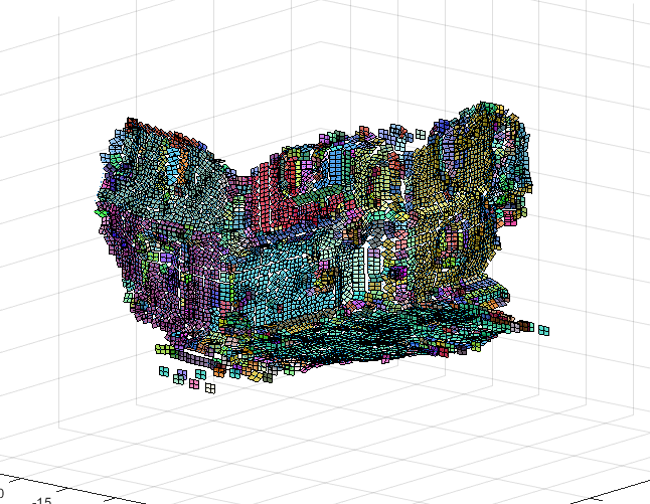}
		\caption{ }
		\label{fig:geostruct pc rg}
	\end{subfigure}
	\begin{subfigure}[b]{0.23\textwidth}
		\includegraphics[width=\textwidth]{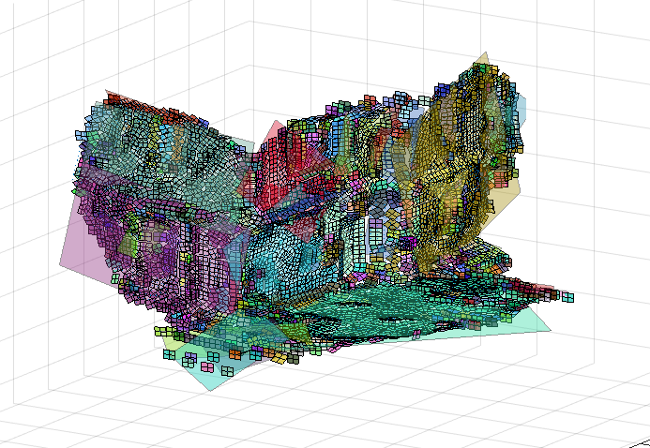}
		\caption{ }
		\label{fig:geostruct pc complete}
	\end{subfigure}
	\caption{The full process of the GeoStructure pipeline is illustrated above, with each of the main stages of point cloud operations shown. (a) shows in input point cloud, after a Mahalanobis filter is applied; (b) shows the structure after the voxel fit process; (c) shows the result of the region growing process, with different region denoted by different colours; (d) illustrates post algorithmic completion, with the regions each aggregated into single planes with composite orientations. These macroscale region planes are interactive and provide the user with $P$ measurements, along with their accuracy.}
	\label{fig:geostructure full}
\end{figure*}
\section{Experimental Results}

\subsection{Experiment 1: $\zeta$}
$\zeta$ is a quantity expressing the proportional size of the search voxel compared to the overall size of the point space. As $\zeta$ progresses to higher values, the resolution will decrease.  This may also explain the shape of the time curve and the small time between Region Growing completion and total completion time. There is a sharp increase in the overall quality between the first an second run. This could be the result of interplay between the voxel scaling and the specific orientation of the surfaces, and, is a candidate for further exploration.

\subsection{Experiment 2: $\theta$}
An optimum range for $\theta$ is between 3 and 8, where the Quality Metric sits between $z$~=~0.1 and $z$~=~0.4. After this, it rises steeply to $z$~=~0.8 and then on to $z$~=~1, which signifies the maximum level of relative inaccuracy.

\subsection{Experiment 3: $\psi$}
Examining the output from the algorithm, a clear optimum range is discernible between $\psi$~=~0.01 and $\psi$~=~0.2 where the overall quality is between $z$~=~0.05 and $z$~=~0.5. After this it rises rapidly to $z$~=~0.6 and $z$~=~0.8. As $\psi$~=~0.4, this is far enough from any seed plane that all candidate planes start to become more difficult to distinguish.

\subsection{Experiment 4: $k$}
Examining the output from the algorithm, a clear optimum point is between $k$~=~3 and $k$~=~11 where the overall quality is just above $z$~=~0.1. After this it rises to $z$~=~0.3 and plateaus there. A smaller $k$ allows for better differentiation between planes, helping to ensure that during region growing, candidate points do not ``jump" over smaller delineating features and grow regions in incorrect areas.

\subsection{Overall Evaluation}

In all cases more optimal results (classified here as $z$ values between 0 and 0.4) occur in almost universally at lower values for each Variable Factor. Also, all Factors, with the exception of $\zeta$, start at a higher $z$ before dropping to a much lower $z$, then rising again (usually at a more gradual interval), suggesting the ranges where optimum values for $z$ sit. 

Generally the relationship between the Region Growing completion and algorithm completion is uniform, with the exception of $\zeta$. There a much greater time elapses for low values, than high values and the difference between curves for Region Growing and algorithm completion are much less. 

Taking these individual results together, it can be concluded that lower values for each of the variables are more optimal, with the resultant increased time for GeoStructure processing being approximately 500~sec.

\section{Conclusion and Future Work}
An algorithm for automated extraction of Planar Orientations from software-generated point clouds has been demonstrated. The algorithm consists of several sub-routines and techniques to automatically manage and process various aspects of the process in taking raw data samples and producing a visual output, that provides $P$ information to the \textit{non-technical user}, with a specified level of accuracy. 

The preprocessing consists of creation of the point clouds from images framed from video as 30 fps, using SfM software (VSfM and PMVS). The point cloud was smoothed and tessellated using Poisson Surface Reconstruction, ready to be passed to the pipeline: GeoStructure. GeoStructure, utilises Mahalanobis distance to remove outlier points, $k$NN region growing using four Variable Factors to segment the point cloud. Accuracy to ground truth is determined using a Quality Metric ($z$) which is dimensionless $\in~[0, 1]$. 

Experiments were conducted to demonstrate ranges of optimum values for each factor and that varying these can improve the performance in accuracy between detected and ground-truthed values. The algorithm produced a final data product that could be directly interacted with by the user (Fig.~\ref{fig:geostruct pc complete}).

Future work involves the application of the algorithm to point clouds derived from real geological structures, and further refinement to the sub-routines of GeoStructure to improve accuracy and visual output representation, so that significant structure is more easily discernable for the non-technical user.

\section*{Declaration of Competing Interest}
The authors declare that they have no known competing financial interests or personal relationships that could have appeared to influence the work reported in this paper.

\section*{Code Availability}
Details of the work described in this text as "the algorithm", "software pipeline", "pipeline", "GeoStructure" are given below:
\\Name: GeoStructure
\\Developer: J. Kissi
\\Available: 2020
\\Address: Department of Electrical and Computer Engineering, Thompson Engineering Building, Western University, London, Ontario, Canada, N6A 5B9
\\email: jkissiam@uwo.ca
\\HW: Runs on a PC with a 2.3 GHz processor, minimum of 8Gb RAM 
\\SW: Minimum MATLAB 2016 is required; full installation, with all packages preferred (not mandatory)
\\Size: Under 1Gb HDD space required to run
\\ SW pipeline is available @ 
\\https://github.com/jkissi/GeoStructure



\normalsize
\bibliography{literary}

\begin{thebibliography}{51}
\providecommand{\natexlab}[1]{#1}
\providecommand{\url}[1]{\texttt{#1}}
\expandafter\ifx\csname urlstyle\endcsname\relax
  \providecommand{\doi}[1]{doi: #1}\else
  \providecommand{\doi}{doi: \begingroup \urlstyle{rm}\Url}\fi

\bibitem[Fosson(2016)]{Fosson2016}
Haakon Fosson.
\newblock \emph{{Structural Geology}}.
\newblock Cambridge University Press, 2 edition, 2016.
\newblock ISBN 978-1-10705764-7.

\bibitem[Sirat and Talbot(2001)]{Sirat2001}
M.~Sirat and C.~J. Talbot.
\newblock {Application of artificial neural networks to fracture analysis at
  the ??sp?? HRL, Sweden: Fracture sets classification}.
\newblock \emph{International Journal of Rock Mechanics and Mining Sciences},
  38\penalty0 (5):\penalty0 621--639, 2001.
\newblock ISSN 13651609.
\newblock \doi{10.1016/S1365-1609(01)00030-2}.

\bibitem[Feng et~al.(2001)Feng, Sj{\"{o}}gren, Stephansson, and Jing]{Feng2001}
Q.~Feng, P.~Sj{\"{o}}gren, O.~Stephansson, and L.~Jing.
\newblock {Measuring fracture orientation at exposed rock faces by using a
  non-reflector total station}.
\newblock \emph{Engineering Geology}, 59\penalty0 (1-2):\penalty0 133--146,
  2001.
\newblock ISSN 00137952.
\newblock \doi{10.1016/S0013-7952(00)00070-3}.

\bibitem[Post et~al.(2001)Post, Kemeny, and Murphy]{Post2001}
R~M Post, J~M Kemeny, and R~Murphy.
\newblock {Image processing for automatic extraction of rock joint orientation
  data from digital images a forward}.
\newblock \emph{Rock Mechanics in the National Interest}, 2001.

\bibitem[Kemeny and Post(2003)]{Kemeny2003}
John Kemeny and Randy Post.
\newblock {Estimating three-dimensional rock discontinuity orientation from
  digital images of fracture traces}.
\newblock \emph{Computers and Geosciences}, 29\penalty0 (1):\penalty0 65--77,
  2003.
\newblock ISSN 00983004.
\newblock \doi{10.1016/S0098-3004(02)00106-1}.

\bibitem[Olaniyan et~al.(2014)Olaniyan, Smith, Lafrance, and
  Pinet]{Olaniyan2014}
Oladele Olaniyan, Richard~S. Smith, Bruno Lafrance, and Nicolas Pinet.
\newblock {A constrained potential field data interpretation of the deep
  geometry of the Sudbury structure}.
\newblock \emph{Canadian Journal of Earth Sciences}, 51\penalty0 (7):\penalty0
  715--729, 2014.
\newblock ISSN 0008-4077.
\newblock \doi{10.1139/cjes-2013-0212}.
\newblock URL
  \url{http://www.nrcresearchpress.com/doi/abs/10.1139/cjes-2013-0212}.

\bibitem[Adam et~al.(2000)Adam, Perron, Milkereit, Wu, Calvert, Salisbury,
  Verpaelst, and Dion]{Adam2000}
Erick Adam, G~Perron, B~Milkereit, Jianjun Wu, A~J Calvert, M~Salisbury, Pierre
  Verpaelst, and Denis-Jacques Dion.
\newblock {A review of high-resolution seismic profiling across the Sudbury,
  Selbaie, Noranda, and Matagami mining camps}.
\newblock \emph{Canadian Journal of Earth Sciences}, 37\penalty0
  (2-3):\penalty0 503--516, 2000.
\newblock ISSN 0008-4077.
\newblock \doi{10.1139/e99-064}.

\bibitem[Hecht et~al.(2008)Hecht, Wittek, Riller, Mohr, Schmitt, and
  Grieve]{Hecht2008}
Lutz Hecht, A~Wittek, Ulrich Riller, T~Mohr, Ralf~Thomas Schmitt, and Richard
  A.~F. Grieve.
\newblock {Differentiation and emplacement of the Worthington Offset Dike of
  the Sudbury impact structure, Ontario}.
\newblock \emph{Meteoritics {\&} Planetary Science}, 43\penalty0 (10):\penalty0
  1659--1679, 2008.
\newblock ISSN 10869379.
\newblock \doi{10.1111/j.1945-5100.2008.tb00635.x}.

\bibitem[Chen et~al.(2015)Chen, Ni, Kapp, Chen, Xiao, and Li]{Chen2015}
Ninghua Chen, Nina Ni, Paul Kapp, Jianyu Chen, Ancheng Xiao, and Hongge Li.
\newblock {Structural Analysis of the Hero Range in the Qaidam Basin,
  Northwestern China, Using Integrated UAV, Terrestrial LiDAR, Landsat 8, and
  3-D Seismic Data}.
\newblock \emph{IEEE Journal of Selected Topics in Applied Earth Observations
  and Remote Sensing}, 8\penalty0 (9):\penalty0 4581--4591, 2015.

\bibitem[Rousell et~al.(2003)Rousell, Fedorowich, and Dressler]{Rousell2003a}
Don~H. Rousell, John~S. Fedorowich, and Burkhard~O. Dressler.
\newblock {Sudbury Breccia (Canada): A product of the 1850 Ma Sudbury Event and
  host to footwall Cu-Ni-PGE deposits}.
\newblock \emph{Earth-Science Reviews}, 60\penalty0 (3-4):\penalty0 147--174,
  2003.
\newblock ISSN 00128252.
\newblock \doi{10.1016/S0012-8252(02)00091-0}.

\bibitem[Tuchscherer and Spray(2002)]{Tuchscherer2002a}
M.~G. Tuchscherer and J.~G. Spray.
\newblock {Geology, mineralization, and emplacement of the Foy Offset Dike,
  Sudbury impact structure}.
\newblock \emph{Economic Geology}, 97\penalty0 (7):\penalty0 1377--1397, 2002.
\newblock ISSN 03610128.
\newblock \doi{10.2113/gsecongeo.97.7.1377}.

\bibitem[Boerner et~al.(2000)Boerner, Kurtz, and Craven]{Boerner2000a}
David~E Boerner, Ron~D Kurtz, and James~a Craven.
\newblock {A summary of electromagnetic studies on the Abitibi–Grenville
  transect}.
\newblock \emph{Canadian Journal of Earth Sciences}, 437:\penalty0 427--437,
  2000.

\bibitem[Wood et~al.(1998)Wood, Wood, Spray, and Spray]{Wood1998}
Christina~R Wood, Christina~R Wood, John~G Spray, and John~G Spray.
\newblock {Origin and emplacement of offset dykes in the Sudbury impact
  structure: Constraints from Hess}.
\newblock \emph{Meteoritics {\&} Planetary Science}, 33\penalty0 (2):\penalty0
  337--347, 1998.
\newblock ISSN 1086-9379.
\newblock \doi{10.1111/j.1945-5100.1998.tb01638.x}.
\newblock URL
  \url{http://onlinelibrary.wiley.com/doi/10.1111/j.1945-5100.1998.tb01638.x/abstract{\%}5Cnpapers3://publication/uuid/64ADFD6C-5741-4222-B186-C35491144245}.

\bibitem[Haid(2016)]{Haid2016}
Taylor~M Haid.
\newblock \emph{{Utilization of Lidar Intensity Data and Passive Visible
  Imagery for Geological Mapping of Planetary Surfaces}}.
\newblock PhD thesis, University of Western Ontario, 2016.
\newblock URL \url{https://ir.lib.uwo.ca/etd/4187}.

\bibitem[Mah et~al.(2013)Mah, Samson, McKinnon, and Thibodeau]{Mah2013}
Jason Mah, Claire Samson, Stephen~D. McKinnon, and Denis Thibodeau.
\newblock {3D laser imaging for surface roughness analysis}.
\newblock \emph{International Journal of Rock Mechanics and Mining Sciences},
  58:\penalty0 111--117, 2013.
\newblock ISSN 13651609.
\newblock \doi{10.1016/j.ijrmms.2012.08.001}.
\newblock URL \url{http://dx.doi.org/10.1016/j.ijrmms.2012.08.001}.

\bibitem[Vasuki et~al.(2014)Vasuki, Holden, Kovesi, and
  Micklethwaite]{Vasuki2014}
Yathunanthan Vasuki, Eun~Jung Holden, Peter Kovesi, and Steven Micklethwaite.
\newblock {Semi-automatic mapping of geological Structures using UAV-based
  photogrammetric data: An image analysis approach}.
\newblock \emph{Computers and Geosciences}, 69:\penalty0 22--32, 2014.
\newblock ISSN 00983004.
\newblock \doi{10.1016/j.cageo.2014.04.012}.
\newblock URL \url{http://dx.doi.org/10.1016/j.cageo.2014.04.012}.

\bibitem[Bellian(2005)]{Bellian2005}
J.A. Bellian.
\newblock {Digital Outcrop Models: Applications of Terrestrial Scanning Lidar
  Technology in Stratigraphic Modeling}.
\newblock \emph{Journal of Sedimentary Research}, 75\penalty0 (2):\penalty0
  166--176, 2005.
\newblock ISSN 1527-1404.
\newblock \doi{10.2110/jsr.2005.013}.

\bibitem[Tavani et~al.(2014)Tavani, Granado, Corradetti, Girundo, Iannace,
  Arbu{\'{e}}s, Mu{\~{n}}oz, and Mazzoli]{Tavani2014}
S.~Tavani, P.~Granado, A.~Corradetti, M.~Girundo, A.~Iannace, P.~Arbu{\'{e}}s,
  J.~A. Mu{\~{n}}oz, and S.~Mazzoli.
\newblock {Building a virtual outcrop, extracting geological information from
  it, and sharing the results in Google Earth via OpenPlot and Photoscan: An
  example from the Khaviz Anticline (Iran)}.
\newblock \emph{Computers and Geosciences}, 63:\penalty0 44--53, 2014.
\newblock ISSN 00983004.
\newblock \doi{10.1016/j.cageo.2013.10.013}.
\newblock URL \url{http://dx.doi.org/10.1016/j.cageo.2013.10.013}.

\bibitem[Vosselman et~al.(2004{\natexlab{a}})Vosselman, Gorte, Sithole, and
  Rabbani]{Vosselman2004a}
G.~Vosselman, B~G~H Gorte, G~Sithole, and T.~Rabbani.
\newblock {Recognising Structure in Laser Scanner Point Clouds}.
\newblock \emph{The International Archives of the Photogrammetry, Remote
  Sensing and Spatial Information Sciences}, pages 1--6, 2004{\natexlab{a}}.

\bibitem[Brown and Lowe(2002)]{Brown2002}
M~Brown and D~G Lowe.
\newblock {Invariant Features from Interest Point Groups}.
\newblock \emph{British Machine Vision Conference, BMVC 2002}, pages 656--665,
  2002.
\newblock \doi{10.1.1.1.8475}.
\newblock URL \url{http://www.bmva.org/bmvc/2002/index.html}.

\bibitem[Brown and Lowe(2005)]{Brown2005}
M.~Brown and D.~G. Lowe.
\newblock {Unsupervised 3D object recognition and reconstruction in unordered
  datasets}.
\newblock \emph{Proceedings of International Conference on 3-D Digital Imaging
  and Modeling, 3DIM}, pages 56--63, 2005.
\newblock ISSN 15506185.
\newblock \doi{10.1109/3DIM.2005.81}.

\bibitem[Lowe(1999{\natexlab{a}})]{Lowe1999a}
David~G. Lowe.
\newblock {Object recognition from local scale-invariant features}.
\newblock \emph{Proceedings of the IEEE International Conference on Computer
  Vision}, 2:\penalty0 1150--1157, 1999{\natexlab{a}}.
\newblock \doi{10.1109/iccv.1999.790410}.

\bibitem[Lowe(2004)]{Lowe2004}
D.~G. Lowe.
\newblock {Distinctive image features from scale invariant keypoints}.
\newblock \emph{Int'l Journal of Computer Vision}, 60\penalty0 (2):\penalty0
  91--110, 2004.
\newblock ISSN 0920-5691.
\newblock \doi{http://dx.doi.org/10.1023/B:VISI.0000029664.99615.94}.
\newblock URL \url{http://portal.acm.org/citation.cfm?id=996342}.

\bibitem[Gigli and Casagli(2011)]{Gigli2011}
Giovanni Gigli and Nicola Casagli.
\newblock {Semi-automatic extraction of rock mass structural data from high
  resolution LIDAR point clouds}.
\newblock \emph{International Journal of Rock Mechanics and Mining Sciences},
  48\penalty0 (2):\penalty0 187--198, 2011.
\newblock ISSN 13651609.
\newblock \doi{10.1016/j.ijrmms.2010.11.009}.
\newblock URL \url{http://dx.doi.org/10.1016/j.ijrmms.2010.11.009}.

\bibitem[Westoby et~al.(2012)Westoby, Brasington, Glasser, Hambrey, and
  Reynolds]{Westoby2012}
M.~J. Westoby, J.~Brasington, N.~F. Glasser, M.~J. Hambrey, and J.~M. Reynolds.
\newblock {`Structure-from-Motion' photogrammetry: A low-cost, effective tool
  for geoscience applications}.
\newblock \emph{Geomorphology}, 179:\penalty0 300--314, 2012.
\newblock ISSN 0169555X.
\newblock \doi{10.1016/j.geomorph.2012.08.021}.
\newblock URL \url{http://dx.doi.org/10.1016/j.geomorph.2012.08.021}.

\bibitem[Hartley and Zisserman(2003)]{Hartley2003}
Richard Hartley and Andrew Zisserman.
\newblock \emph{{Multiple View Geometry in Computer Vision}}.
\newblock Cambridge University Press, 2nd edition, 2003.
\newblock ISBN 978-0521540513.

\bibitem[Harwin and Lucieer(2012)]{Harwin2012}
Steve Harwin and Arko Lucieer.
\newblock {Assessing the accuracy of georeferenced point clouds produced via
  multi-view stereopsis from Unmanned Aerial Vehicle (UAV) imagery}.
\newblock \emph{Remote Sensing}, 4\penalty0 (6):\penalty0 1573--1599, 2012.
\newblock ISSN 20724292.
\newblock \doi{10.3390/rs4061573}.

\bibitem[James and Robson(2012)]{James2012}
M.~R. James and S.~Robson.
\newblock {Straightforward reconstruction of 3D surfaces and topography with a
  camera: Accuracy and geoscience application}.
\newblock \emph{Journal of Geophysical Research: Earth Surface}, 117\penalty0
  (3):\penalty0 1--17, 2012.
\newblock ISSN 21699011.
\newblock \doi{10.1029/2011JF002289}.

\bibitem[Dall'Asta et~al.(2015)Dall'Asta, Thoeni, Santise, Forlani, Giacomini,
  and Roncella]{DallAsta2015}
Elisa Dall'Asta, Klaus Thoeni, Marina Santise, Gianfranco Forlani, Anna
  Giacomini, and Riccardo Roncella.
\newblock {Network Design and Quality Checks in Automatic Orientation of
  Close-Range Photogrammetric Blocks}.
\newblock \emph{Sensors}, 15\penalty0 (4):\penalty0 7985--8008, 2015.
\newblock ISSN 1424-8220.
\newblock \doi{10.3390/s150407985}.
\newblock URL \url{http://www.mdpi.com/1424-8220/15/4/7985/}.

\bibitem[James and Robson(2014)]{James2014}
Mike~R James and Stuart Robson.
\newblock {Mitigating systematic error in topographic models derived from UAV
  and ground-based image networks}.
\newblock \emph{Earth Surface Processes and Landforms}, 39\penalty0
  (10):\penalty0 1413--1420, 2014.
\newblock ISSN 10969837.
\newblock \doi{10.1002/esp.3609}.

\bibitem[Bemis et~al.(2014)Bemis, Micklethwaite, Turner, James, Akciz, Thiele,
  and Ali]{Bemis2014}
Sean~P Bemis, Steven Micklethwaite, Darren Turner, Mike~R James, Sinan Akciz,
  Sam~T Thiele, and Hasnain Ali.
\newblock {Ground-based and UAV-Based photogrammetry: A multi-scale,
  high-resolution mapping tool for structural geology and paleoseismology}.
\newblock \emph{Journal of Structural Geology}, 69:\penalty0 163--178, 2014.
\newblock ISSN 0191-8141.
\newblock \doi{10.1016/j.jsg.2014.10.007}.
\newblock URL \url{http://dx.doi.org/10.1016/j.jsg.2014.10.007}.

\bibitem[Rabbani et~al.(2006)Rabbani, van~den Heuvel, Vosselman, van~den
  Wildenberg, and Vosselman]{Rabbani2006}
T.~Rabbani, F~a van~den Heuvel, G.~Vosselman, F.~van~den Wildenberg, and
  G.~Vosselman.
\newblock {Segmentation of point clouds using smoothness constraint}.
\newblock \emph{International Archives of Photogrammetry, Remote Sensing and
  Spatial Information Sciences}, 36\penalty0 (5):\penalty0 248--253, 2006.
\newblock ISSN 16821750.
\newblock URL
  \url{http://www.isprs.org/proceedings/XXXVI/part5/paper/RABB{\_}639.pdf}.

\bibitem[Xi et~al.(2009)Xi, Duan, and Zhao]{Xi2009}
Yongjian Xi~Yongjian Xi, Ye~Duan~Ye Duan, and Hongkai Zhao~Hongkai Zhao.
\newblock {A nonparametric approach for noisy point data preprocessing}.
\newblock \emph{2009 11th IEEE International Conference on Computer-Aided
  Design and Computer Graphics}, 1\penalty0 (2):\penalty0 2--7, 2009.
\newblock \doi{10.1109/CADCG.2009.5246900}.

\bibitem[Brophy(2015)]{Brophy}
M~A Brophy.
\newblock {Surface Reconstruction from Noisy and Sparse Data}.
\newblock \emph{UWO Electronic Thesis and Dissertation Repository}, Paper 3375,
  2015.

\bibitem[Guerrero et~al.(2018)Guerrero, Kleiman, Ovsjanikov, and
  Mitra]{Guerrero2018}
Paul Guerrero, Yanir Kleiman, Maks Ovsjanikov, and Niloy~J. Mitra.
\newblock {PCPNet learning local shape properties from raw point clouds}.
\newblock \emph{Computer Graphics Forum}, 37\penalty0 (2):\penalty0 75--85,
  2018.
\newblock ISSN 14678659.
\newblock \doi{10.1111/cgf.13343}.

\bibitem[Kissi(2016)]{Kissi2016}
Jon Kissi.
\newblock \emph{{Automatic Fracture Orientation Extraction from SfM Point
  Clouds}}.
\newblock Mesc, University of Western Ontario, 2016.
\newblock URL \url{https://ir.lib.uwo.ca/etd/4243}.

\bibitem[Lowe(1999{\natexlab{b}})]{Lowe1999}
D.G.~G Lowe.
\newblock {Object Recognition fromLocal Scale-Invariant Features}.
\newblock \emph{IEEE International Conference on Computer Vision},
  1999{\natexlab{b}}.
\newblock ISSN 0-7695-0164-8.
\newblock \doi{10.1109/ICCV.1999.790410}.

\bibitem[Vosselman et~al.(2004{\natexlab{b}})Vosselman, Gorte, Sithole, and
  Rabbani]{Vosselman2004}
George Vosselman, B~G~H Gorte, G~Sithole, and Tahir Rabbani.
\newblock {Recognising strucutre in laser scanner point clouds}.
\newblock \emph{Remote Sensing and Spatial Information Sciences}, pages 33--38,
  2004{\natexlab{b}}.
\newblock ISSN 00063525.
\newblock \doi{10.1002/bip.360320508}.

\bibitem[Wang and Lu(2009)]{Wang2009}
Chi~Kuei Wang and Yao~Yu Lu.
\newblock {Potential of ILRIS3D intensity data for planar surfaces
  segmentation}.
\newblock \emph{Sensors}, 9\penalty0 (7):\penalty0 5770--5782, 2009.
\newblock ISSN 14248220.
\newblock \doi{10.3390/s90705770}.

\bibitem[McLeod(2012)]{McLeod2013}
Tara McLeod.
\newblock \emph{{Three-dimensional imaging applications in Earth Sciences using
  video data acquired from an unmanned aerial vehicle}}.
\newblock Doctoral dissertation, Carleton University, 2012.
\newblock URL
  \url{http://ezproxy.net.ucf.edu/login?url=http://search.proquest.com/docview/1418482332?accountid=10003{\%}5Cnhttp://sfx.fcla.edu/ucf?url{\_}ver=Z39.88-2004{\&}rft{\_}val{\_}fmt=info:ofi/fmt:kev:mtx:dissertation{\&}genre=dissertations+{\&}+theses{\&}sid=ProQ:ProQuest+Dissertations+{\&}+}.

\bibitem[Wu et~al.(2011)Wu, Agarwal, Curless, and Seitz]{Wu2011}
C.~Wu, S.~Agarwal, B.~Curless, and S.~M. Seitz.
\newblock {Multicore bundle adjustment}.
\newblock \emph{Computer Vision and Pattern Recognition (CVPR)}, pages
  3057--3064, 2011.

\bibitem[Wu(2013)]{Wu2013}
Changchang Wu.
\newblock {Towards linear-time incremental structure from motion}.
\newblock \emph{Proceedings - 2013 International Conference on 3D Vision, 3DV
  2013}, pages 127--134, 2013.
\newblock \doi{10.1109/3DV.2013.25}.

\bibitem[Kazhdan and Hoppe(2006)]{Kazhdan2006}
Michael Kazhdan and Hugues Hoppe.
\newblock {Poisson surface reconstruction}.
\newblock \emph{Proceedings of the fourth Eurographics symposium on Geometry
  processing}, 7, 2006.
\newblock URL
  \url{http://dl.acm.org/citation.cfm?id=2487237{\%}5Cnhttp://dl.acm.org/citation.cfm?doid=2487228.2487237}.

\bibitem[Kazhdan and Hoppe(2013)]{Kazhdan2013a}
Michael Kazhdan and Hugues Hoppe.
\newblock {Screened poisson surface reconstruction}.
\newblock \emph{ACM Transactions on Graphics}, 32\penalty0 (3):\penalty0 1--13,
  2013.
\newblock ISSN 07300301.
\newblock \doi{10.1145/2487228.2487237}.
\newblock URL
  \url{http://dl.acm.org/citation.cfm?id=2487237{\%}5Cnhttp://dl.acm.org/citation.cfm?doid=2487228.2487237}.

\bibitem[Calakli and Taubin(2011)]{Calakli2011}
F~Calakli and G~Taubin.
\newblock {SSD: Smooth signed distance surface reconstruction}.
\newblock \emph{Computer Graphics Forum}, 30\penalty0 (7):\penalty0 1993--2002,
  2011.
\newblock ISSN 14678659.
\newblock \doi{10.1111/j.1467-8659.2011.02058.x}.

\bibitem[Furukawa and Ponce(2010)]{Furukawa2010a}
Yasutaka Furukawa and Jean Ponce.
\newblock {Accurate, dense, and robust multiview stereopsis}.
\newblock \emph{IEEE Transactions on Pattern Analysis and Machine
  Intelligence}, 32\penalty0 (8):\penalty0 1362--1376, 2010.
\newblock ISSN 01628828.
\newblock \doi{10.1109/TPAMI.2009.161}.

\bibitem[Snavely et~al.(2006)Snavely, Seitz, and Szeliski]{Snavely2006}
Noah Snavely, Steven~M Seitz, and Richard Szeliski.
\newblock {Photo tourism: Exploring Photo Collections in 3D}.
\newblock \emph{ACM Transactions on Graphics}, 25\penalty0 (3):\penalty0
  835--846, 2006.
\newblock ISSN 07300301.
\newblock \doi{10.1145/1141911.1141964}.
\newblock URL
  \url{http://doi.acm.org/10.1145/1141911.1141964{\%}5Cnhttp://portal.acm.org/citation.cfm?doid=1141911.1141964}.

\bibitem[Gomez(2014)]{Gomez2014}
Christopher Gomez.
\newblock {Multi-scale Voxel-based Algorithm for UAV-derived Point-clouds of
  Complex Surfaces}.
\newblock pages 205--209, 2014.

\bibitem[Turner and Zakhor(2013)]{Turner2013}
Eric Turner and Avideh Zakhor.
\newblock {Watertight planar surface meshing of indoor point-clouds with voxel
  carving}.
\newblock \emph{Proceedings of the 2013 IEEE International Conference on 3D
  Vision, 3DV 2013}, pages 41--48, 2013.
\newblock ISSN 1550-6185.
\newblock \doi{10.1109/3DV.2013.14}.

\bibitem[Mah et~al.(2011)Mah, Samson, and McKinnon]{Mah2011}
Jason Mah, Claire Samson, and Stephen~D. McKinnon.
\newblock {3D laser imaging for joint orientation analysis}.
\newblock \emph{International Journal of Rock Mechanics and Mining Sciences},
  48\penalty0 (6):\penalty0 932--941, 2011.
\newblock ISSN 13651609.
\newblock \doi{10.1016/j.ijrmms.2011.04.010}.
\newblock URL \url{http://dx.doi.org/10.1016/j.ijrmms.2011.04.010}.

\bibitem[Seitz et~al.(2006)Seitz, Curless, Diebel, Scharstein, and
  Szeliski]{Seitz2006}
S~M Seitz, B~Curless, J~Diebel, D~Scharstein, and R~Szeliski.
\newblock {A Comparison and Evaluation of Multi-View Stereo Reconstruction
  Algorithms}.
\newblock \emph{2006 IEEE Computer Society Conference on Computer Vision and
  Pattern Recognition - Volume 1 (CVPR'06)}, 1:\penalty0 519--528, 2006.
\newblock ISSN 1063-6919.
\newblock \doi{10.1109/CVPR.2006.19}.
\newblock URL
  \url{http://ieeexplore.ieee.org/lpdocs/epic03/wrapper.htm?arnumber=1640800}.

\end{thebibliography}


\end{document}